\pdfoutput=1

\documentclass[11pt]{article}

\usepackage[preprint]{acl}

\usepackage{times}
\usepackage{latexsym}

\usepackage[T1]{fontenc}

\usepackage[utf8]{inputenc}

\usepackage{microtype}

\usepackage{inconsolata}

\usepackage{graphicx}
\usepackage{multirow}

\usepackage{booktabs}
\usepackage{amsfonts}
\usepackage{amsmath}
\usepackage{tcolorbox}
\usepackage{colortbl,array}
\usepackage{enumitem}
\usepackage{xcolor} 

\makeatother
\newcommand{\mytextbox}[2]{\tikzmarknode[draw=#1,thick,inner sep=2pt]{test}{\myfontsize #2}}
\definecolor{myred}{rgb}{0.7, 0.3, 0.0}
\definecolor{myblue}{HTML}{054488}
\definecolor{mygreen}{HTML}{056b34}
\definecolor{myorange}{HTML}{ff8800}
\definecolor{mypurple}{HTML}{8400ff}
\definecolor{mypink}{HTML}{f7acb9}

\newcommand{\red}[1]{\mytextbox{myred}{\textbf{\textcolor{myred}{#1}}}}
\newcommand{\blue}[1]{\mytextbox{myblue}{\textbf{\textcolor{myblue}{#1}}}}
\newcommand{\green}[1]{\mytextbox{mygreen}{\textbf{\textcolor{mygreen}{#1}}}}

\newcommand{\purple}[1]{\mytextbox{mypurple}{\textbf{\textcolor{mypurple}{#1}}}}

\usepackage{booktabs}
\usepackage[table]{xcolor}
\usepackage{tikz}
\usetikzlibrary{tikzmark}
\usepackage{listings}

\newcommand{\myfontsize}{\fontsize{9pt}{11pt}\selectfont}

\title{CAREAgent: Clinical Agent with Structured Reasoning and Tool-Integrated for Order Generation}

\author{
\textbf{Ruihui Hou\textsuperscript{\rm $\diamondsuit$}},
\textbf{Ziyue Huai\textsuperscript{\rm $\diamondsuit$}},
\textbf{Chennuo Zhang\textsuperscript{\rm $\diamondsuit$}},
\textbf{Ziyan Liu\textsuperscript{\rm $\diamondsuit$}},
\textbf{Siran Zhao\textsuperscript{\rm $\diamondsuit$}},\\
\textbf{Yao Yu\textsuperscript{\rm $\clubsuit$}},
\textbf{Jie Zhai\textsuperscript{\rm $\diamondsuit$}\thanks{Corresponding authors}},
\textbf{Tong Ruan\textsuperscript{\rm $\diamondsuit$}\footnotemark[1]}
\\
\textsuperscript{\rm $\diamondsuit$}East China University of Science and Technology, Shanghai, China,
\\
\textsuperscript{\rm $\clubsuit$}Zhongshan Hospital, Fudan University, Shanghai, China.
\\
}

\begin{document}
\maketitle
\begin{abstract}


Clinical order generation serves as a critical bridge between clinical decision-making and real-world practice, translating medical decisions into concrete and executable orders. 
Existing agents mainly focus on coarse-grained decisions and overlook the fine-grained, executable information required for clinical orders.
To address this gap, we propose CAREAgent, an agent for clinical order generation.
To support its training, we introduce a two-stage agentic reasoning data construction method. 
First, we design an agent framework that constructs verifiable reasoning trajectories aligned with realistic clinical tool usage.
Second, we filter reasoning trajectories by format compliance, order validity, and clinical plausibility.
Building on the constructed data, the model is first trained via supervised fine-tuning to acquire fundamental reasoning formats and medical knowledge, and is subsequently optimized through reinforcement learning with multi-dimensional reward functions to enhance complex clinical reasoning capabilities.
Experiments on multiple benchmarks demonstrate the effectiveness of CAREAgent. On ClinicalBench (unseen during training), CAREAgent improves the F1 score by 5.05\%, 2.09\%, and 0.86\% over the single-agent, multi-agent, and agentic reasoning methods, respectively.
\end{abstract}


\section{Introduction}



Large language models (LLMs) have achieved notable progress in clinical decision-making tasks, including clinical diagnosis~\cite{integrating, medreason} and treatment recommendation~\cite{medplan, liu2025automated}. However, most existing work focuses on coarse-grained clinical guidance, with limited attention to translating medical decisions into concrete and executable clinical orders, such as order names, drug dosages, routes of administration, and frequency~\cite{Uncertainty}. 
Clinical order generation serves as a critical interface between clinical decision-making and real-world clinical execution, and its correctness directly impacts treatment outcomes~\cite{qiu2025quantifying, Musen2021}.




\begin{figure}
    \centering
    \includegraphics[width=1\linewidth]{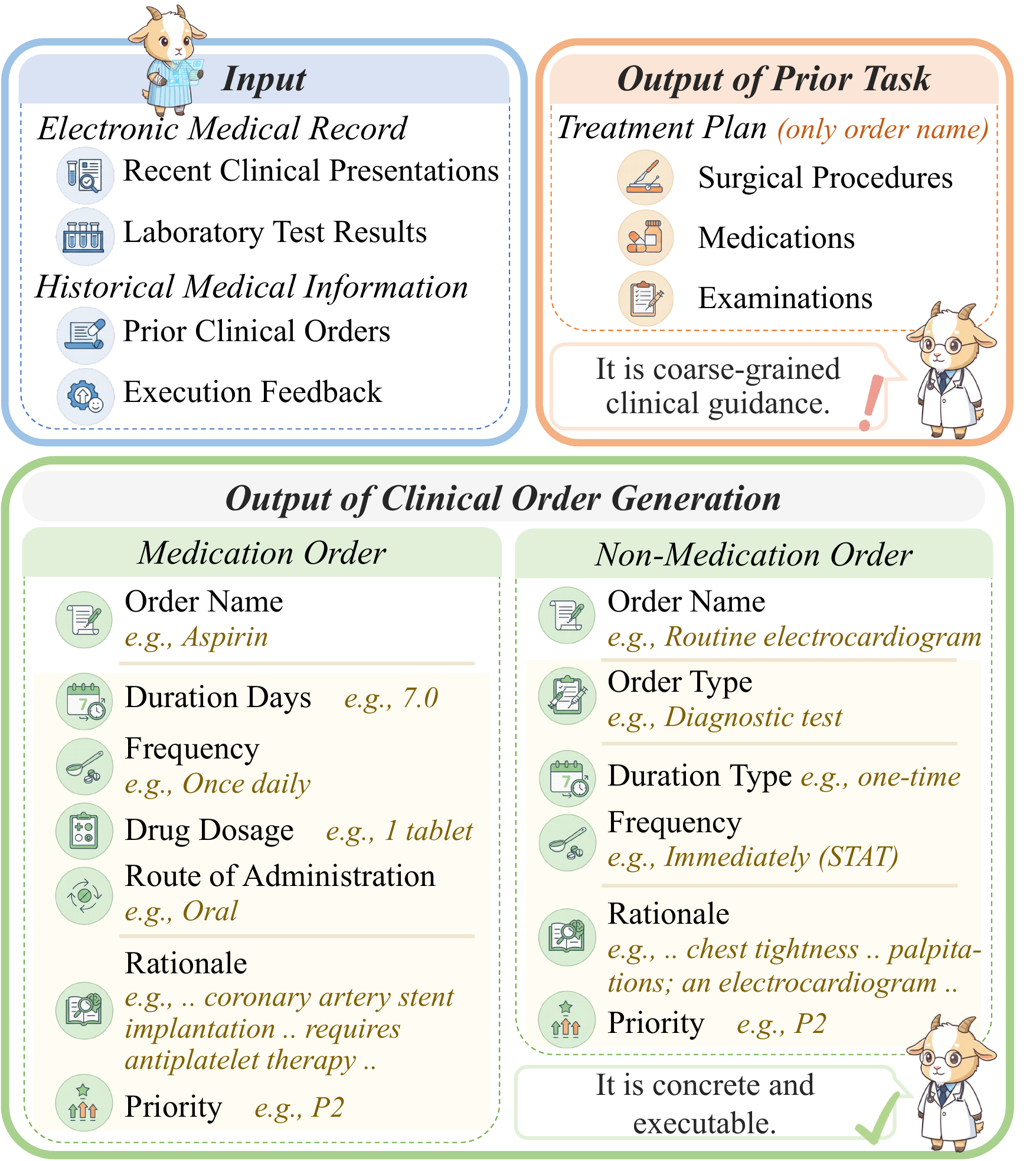}
    \caption{Comparison between clinical order generation and prior treatment recommendation tasks.}
    \label{intro}
    \vspace{-0.2cm}
\end{figure}

The clinical order generation task aims to produce a set of concrete and executable clinical orders, along with appropriate priority rankings, based on patients’ current clinical information (e.g., symptoms and laboratory test results) as well as prior examinations, historical orders, and their execution feedback~\cite{Uncertainty}. 
Each order can be categorized as either a medication or a non-medication order, both of which contain fine-grained attributes such as the order name, rationale, and frequency.
In addition, medication orders specify attributes including duration days, drug dosage, and route of administration, while non-medication orders further include the duration type and order type.
As illustrated in Figure~\ref{intro}, the output of this task is a collection of clinical orders with explicit execution details and priority constraints.
Existing research on medical decision-making can be broadly categorized into three paradigms: single-agent, multi-agent, and agentic reasoning methods~\cite{agentsurvey}. 
Single-agent approaches primarily rely on the medical knowledge and reasoning capabilities of LLMs~\cite{medreason, cod}.
Consequently, they struggle to capture the multi-stage reasoning processes and constraints inherent in real-world clinical decision-making, limiting their ability to generate fine-grained clinical orders.
Multi-agent methods achieve collaborative decision-making through role decomposition and interaction~\cite{Clinicallab, medagents}, which facilitates task decomposition but often depends on manually designed role assignments and coordination protocols, thereby limiting generalization. 
Agentic reasoning approaches model clinical decision-making as a multi-stage process and are commonly trained via cold-start supervised fine-tuning (SFT)~\cite{huatuogpto1} or direct reinforcement learning (RL)~\cite{medresearcher, zheng2025end}. 
However, SFT methods primarily rely on template-based synthetic data, which lack realistic clinical tool invocation processes. 
Furthermore, directly applying RL often fails to achieve fast and stable convergence due to insufficient medical knowledge in the model.


In this paper, we propose \textbf{CAREAgent}, a clinical agent with structured reasoning and tool-integrated for order generation. 
To support its training, we develop a two-stage agentic reasoning data construction method that produces high-quality clinical reasoning trajectories with real tool invocations.
In the first stage, we build a summarization agent that converts incremental daily patient records into summaries.
Building on this, we design an agent framework that integrates structured reasoning templates with real external clinical tool calls, yielding reasoning trajectories that closely follow real-world clinical workflows and are highly verifiable.
In the second stage, we apply a multi-step data filtering process to ensure data quality, including rule-based format validation, LLM–based trajectory evaluation, and order validity checking. 
The filtered samples are divided based on reasoning success: those where the agent successfully completes the reasoning process are used to construct the cold-start SFT dataset, while the remaining challenging samples are set aside for the RL stage.
Under this two-stage training paradigm, the model first acquires fundamental reasoning formats, tool usage conventions, and essential medical knowledge via SFT, and is subsequently optimized with RL using multi-dimensional rewards that promote format compliance, reasoning coherence, and order correctness.
This strategy enables the model to generate clinical order reasoning trajectories that are structurally valid and logically consistent.



Our contributions are summarized as follows:
\begin{itemize}
    \item 
    We design a two-stage agentic reasoning data construction method that generates high-quality clinical reasoning trajectories with real tool invocations, without manual annotation.
    \item 
    We train an order generation agent using a two-stage agentic post-training framework that combines SFT and agentic RL to enable structured reasoning, effective tool usage, and the generation of clinically executable orders.
    \item 
    Extensive experiments on three benchmarks demonstrate the effectiveness of CAREAgent. On ClinicalBench, which is unseen during training, CAREAgent improves the $F_1$ score by 5.05\%, 2.09\%, and 0.86\% over the strongest Single-Agent, Multi-Agent, and Agentic Reasoning baselines, respectively.
\end{itemize}

\section{Related Work}


\subsection{Clinical Order Generation}

The clinical order generation task aims to automatically produce concrete and executable clinical intervention recommendations from patients’ clinical records. 
To improve decision performance, prior work has introduced LLMs into clinical decision tasks~\cite{singhal_large_2023}. 
However, due to limited medical knowledge coverage and delayed knowledge updates, such approaches struggle to handle complex clinical scenarios. 
To address these issues, some studies incorporate external medical knowledge sources (e.g., clinical guidelines) to enhance knowledge grounding~\cite{medplan, clinicalrag, Designing}, while recent work further explores multi-agent collaboration frameworks that leverage role decomposition and information exchange to improve reasoning capability and stability~\cite{MedChain, Clinicallab}.
Nevertheless, these approaches primarily focus on clinical reasoning and information integration, without systematically modeling the structured representation and executability requirements of clinical orders. 
In parallel, another line of research focuses on optimal treatment selection, including decision-making under conflicting outcomes~\cite{Multicriteria}, predictive-factor–based treatment selection~\cite{Selecting}, and optimal treatment assignment strategies~\cite{Optimal}. 
Overall, while LLM-based methods excel at multi-source clinical information understanding and complex reasoning, existing research generally lacks explicit modeling of fine-grained order attributes and order priority ranking, which are core challenges in clinical order generation.


\subsection{Medical Agent Reasoning}


Existing medical agent reasoning can be categorized into three types: single-agent, multi-agent, and agentic reasoning methods~\cite{agentsurvey}.
Single-agent methods primarily rely on the inherent medical knowledge and reasoning capabilities of LLMs, using prompt engineering or limited tool invocation to generate clinical diagnosis~\cite{medreason, MedRAG, cod}, treatment recommendations~\cite{medplan, liu2025automated}, or risk prediction~\cite{Agentmd, menti}. 
However, these methods struggle to model the multi-stage clinical reasoning, leading to inconsistent generation of fine-grained clinical orders with complete fields and explicit execution constraints.
Multi-agent methods address medical tasks by coordinating multiple agents with distinct roles (e.g., surgeons, internists), which interact and collaborate to produce a final decision~\cite{MDAgents, medagents, agentclinic}.
However, these methods often depend on manually designed role assignments and coordination protocols, thereby limiting generalization. 
Agentic reasoning methods typically model clinical problems as multi-stage decision processes, updating patient states during the perception stage, planning and reasoning during the cognition stage, and executing clinical decisions or tool invocations during the action stage.
In terms of training, these methods mainly rely on SFT with end-to-end synthesized data~\cite{huatuogpto1, schick2023toolformer} or direct RL~\cite{zheng2025end, medresearcher}. 
However, end-to-end synthesized data fails to faithfully capture real tool invocation processes, resulting in unreliable execution trajectories.
Moreover, direct RL often leads to slow convergence and unstable training due to insufficient domain knowledge.



\section{Methodology}

\subsection{Task Definition}
Given a patient's Electronic Medical Record (EMR) $D=(E,~H)$, where $E$ denotes the current clinical information (e.g., clinical presentation and laboratory test results) and $H$ represents historical medical information, including prior examinations, executed orders, and their execution feedback.
This task aims to generate a set of subsequent clinical orders $C = \{c_1, c_2, \dots, c_n\}$, where $n$ represents the number of orders.
Each order $c_i$ corresponds to a specific clinical intervention and can be categorized as either a medication or a non-medication order.
A medication order is represented as
$c_i = \{\text{order name}, \text{duration days}, \text{frequency}, \text{drug dosage}, \\ \text{route of administration},  \text{rationale}, \text{priority}\},$
while a non-medication order is represented as
$c_i = \{\text{order name}, \text{order type}, \text{duration type},  \text{frequency}, \\ \text{rationale},  \text{priority}\}.$
As shown in Figure~\ref{intro},  the medication order ``Aspirin'' has a duration of 7.0 days, a frequency of once daily, a dosage of 1 tablet, an oral route of administration, a clinical rationale of ``\dots coronary artery stent implantation \dots requires antiplatelet therapy \dots'', and a priority level of P2.
The non-medication order ``Routine electrocardiogram'' is categorized as a diagnostic test, with a one-time duration type and an immediate frequency. 
Its clinical rationale is ``\dots chest tightness \dots palpitations; an electrocardiogram \dots'', and the priority is P2.


\subsection{Two-stage Agentic Reasoning Data Construction}

\begin{figure*}
    \centering
    \includegraphics[width=1\linewidth]{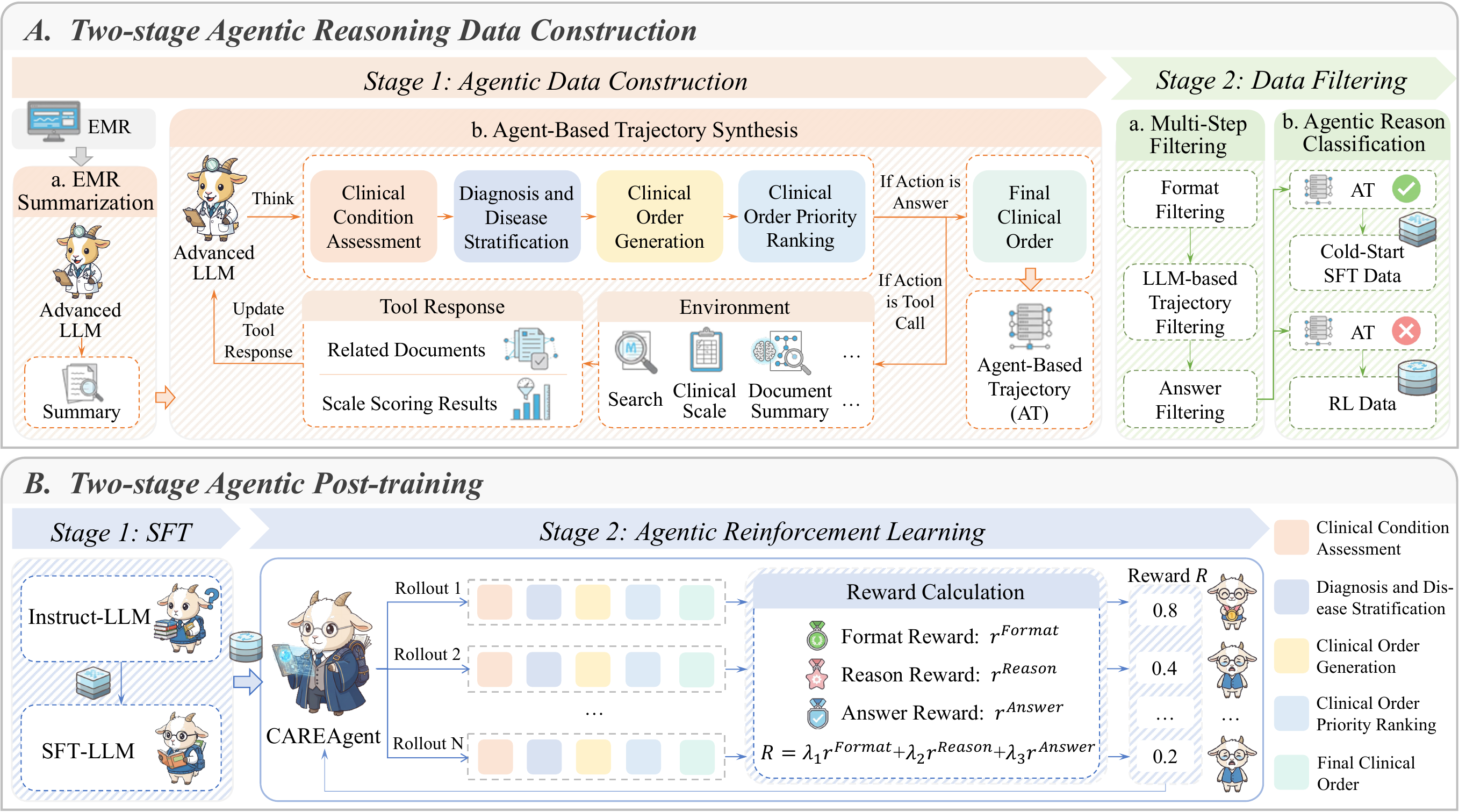}
    \caption{Overview of our framework. The upper part illustrates the two-stage agentic reasoning data construction process, while the lower part presents the two-stage agentic post-training procedure.}
    \label{framework}
\end{figure*}

In this section, we present our pipeline for constructing agentic reasoning data, aiming to enable automated and scalable construction of high-quality tool-use datasets, as illustrated in Figure~\ref{framework}.

\subsubsection{Agentic Data Construction}

During hospitalization, clinicians continuously add daily progress notes, examination results, and treatment records to the EMR. As the length of stay increases, these records accumulate and the overall EMR may exceed the maximum context length that the model can process. To address this issue, we design an EMR summarization agent that incrementally summarizes newly generated daily records into structured representations. These summaries are then used as the primary input for subsequent processing, enabling the system to maintain essential clinical information while keeping the input length manageable.
The detailed summarization workflow is described in Appendix~\ref{summary}.



Based on the summarized EMR, we construct an agent-based interactive environment to collect clinical reasoning trajectories that involve realistic tool usage.
Specifically, the framework proceeds through four sequential stages: clinical condition assessment, diagnosis and disease stratification, clinical order generation, and clinical order priority ranking. 
Within each stage, the LLM performs step-by-step reasoning and may invoke external clinical tools when additional evidence is required. Tool invocations are generated using special tags (e.g., <tool\_call>\dots</tool\_call>), which are parsed by the agent framework to call the corresponding external tools. The returned results are then incorporated into the context and used for subsequent reasoning.
This reasoning–tool interaction may occur multiple times during a stage and continues until a predefined maximum number of steps is reached or the model produces a final answer using the tag <answer>\dots</answer>. 
Through this iterative reasoning and tool interaction process, the framework constructs clinical reasoning trajectories that closely mimic real-world diagnostic workflows.

\subsubsection{Data Filtering and Classification}

To select high-quality reasoning trajectories for SFT and RL, we perform systematic data filtering and selection.
\textbf{(1) Data Filtering}. We employ a three-step filtering pipeline to ensure well-formed reasoning trajectories, valid tool invocations, and accurate final answers. 
Specifically:
(i) \textit{Format Filtering}. We remove samples with structural errors, including duplicated decision stages within a trajectory, consecutive tool calls with identical names and parameters, or malformed control tags (e.g., <thought>, <tool\_call>, <answer>).
(ii) \textit{Answer Filtering}. We use a task-specific prompt to validate clinical order names and discard samples containing invalid or incorrect orders.
(iii) \textit{LLM-based Trajectory Filtering}. 
We further evaluate LLM-generated reasoning trajectories by assessing tool invocation appropriateness and key reasoning steps, including clinical condition assessment, diagnosis and stratification, clinical order generation, and clinical priority ranking.
This pipeline ensures that the final dataset contains only high-quality, concise, and executable reasoning trajectories.
\textbf{(2) Data Classification.} To support the learning paradigm from easy to difficult~\cite{abilities} and enhance the model’s reasoning ability, we partition the filtered data into two categories: data for SFT and data for RL. 
For samples where the agent framework successfully completes the reasoning process, we retain the corresponding trajectories to construct a cold-start SFT dataset, $D_{\mathrm{SFT}}$.
Samples that remain challenging are treated as hard examples and reserved for the RL stage, forming the dataset $D_{\mathrm{RL}}$. 
This design enables the model to acquire fundamental tool-usage skills during cold-start SFT and progressively extend to more complex scenarios (e.g., multi-tool collaboration) during RL, thereby facilitating gradual learning and capability enhancement.

\subsection{Two-stage Agentic Post-training}


In this section, we propose a two-stage collaboration training framework. In the SFT stage, the framework enhances the model’s understanding of format and foundational clinical knowledge. In the RL stage, it further improves the model’s ability to perform complex clinical reasoning.

\subsubsection{Supervised Fine-tuning}


To enable the LLM to learn predefined formats and acquire domain-specific clinical knowledge, we perform SFT on the backbone model $P_\theta$ with parameters $\theta$ using samples $(x_i, y_i) \in D_{\mathrm{SFT}}$. 
The training objective is defined as:
\begin{equation}
\mathcal{L}(\boldsymbol{\theta}) = -\sum_{(x_i, y_i) \in D_{\mathrm{SFT}}} \log P_\theta(y_i \mid x_i),
\end{equation}
where $x_i$ denotes the $i$-th input instance. 
Through this process, we obtain an LLM equipped with fundamental reasoning formats, tool usage conventions, and essential medical knowledge.


\subsubsection{Agentic Reinforcement Learning}


After completing SFT, the model is able to generate clinical reasoning trajectories that include tool calls. 
To further enhance its reasoning ability, we perform RL training on challenging samples to improve the model’s performance in complex clinical scenarios.

\textbf{Reward Design.}
To encourage the model to generate trajectories with correct formatting, sound reasoning, and accurate answers, we design reward functions along three dimensions. 
Specifically, for the $i$-th sample, the metric $r_i^{\mathrm{Format}}$ computes the reward for format correctness, assigning $-1$ to invalid formats and $+1$ to valid formats. 
The metric $r_i^{\mathrm{Reason}}$ evaluates the logical soundness of the reasoning chain by assessing four reasoning steps, assigning $-1$ to logically flawed trajectories and $+1$ to logically correct ones. 
The metric $r_i^{\mathrm{Answer}}$ measures answer accuracy, assigning $-1$ to incorrect answers and $+1$ to correct answers.
The correctness of reasoning and answers is evaluated using task-specific prompts in conjunction with an LLM, with detailed prompts provided in Appendix~\ref{prompt_design}.
Finally, the overall reward $R_i$ is computed as:
\begin{equation}
R_i=\lambda_1 r_i^{\mathrm{Format}} + \lambda_2 r_i^{\mathrm{Reason}} + \lambda_3 r_i^{\mathrm{Answer}},
\end{equation}
where $\lambda_1=0.1$, $\lambda_2=0.3$, and $\lambda_3=0.6$ denote the weighting coefficients for the format, reason chain, and the answer, respectively. 


\textbf{Policy Optimization Objective.}
We employ the Group Relative Policy Optimization (GRPO) algorithm~\cite{deepseekmath} to train the model using mixed final rewards.
Formally, at each training step, given a patient’s EMR $C$, the model generates a set of candidate responses $O = \{O_1, O_2, \ldots, O_G\}$, where $G$ denotes the group size.
Each response comprises two components: (1) a structured reasoning chain and (2) the corresponding tool invocations. 
These responses are then evaluated using reward functions, producing a set of rewards $\{R_1, R_2, \ldots, R_G\}$.
The policy $\pi_{\theta}$ is then optimized by maximizing the clipped GRPO objective:
\vspace{-0.1cm}
\begin{equation}
\begin{aligned}
\mathcal{L}_{\mathrm{GRPO}}(\theta) 
&= \mathbb{E}_{C \sim D_{RL}, \{O_i\}_{i=1}^G \sim \pi_{\theta_{\mathrm{old}}}(O \mid C)} \\
&\Biggl[
  \frac{1}{G} \sum_{i=1}^G
  \min\!\Bigl(
    \frac{\pi_{\theta}(O_i \mid C)}{\pi_{\theta_{\mathrm{old}}}(O_i \mid C)}\,A_i,\, \\
    &\mathrm{clip}\!\Bigl(
      \frac{\pi_{\theta}(O_i \mid C)}{\pi_{\theta_{\mathrm{old}}}(O_i \mid C)},
      1-\varepsilon,\,
      1+\varepsilon
    \Bigr)
    A_i
  \Bigr)  \\
  &-\,\beta\,D_{\mathrm{KL}}\bigl(\pi_{\theta}\,\big\|\,\pi_{\mathrm{ref}}\bigr)
\Biggr],
\end{aligned}
\label{eq1}
\end{equation}
where $\varepsilon$ and $\beta$ are hyperparameters that define the clipping range for policy updates and control the strength of the KL divergence penalty, respectively~\cite{deepseekmath}. 
The KL divergence term constrains the updated policy, preventing excessive deviation from a reference model $\pi_{\mathrm{ref}}$, which typically corresponds to the initial model before RL begins.
$A_i$ represents the relative advantage of the $i$-th response, which is  computed as follows:
\begin{equation}
A_i = \frac{R_i - \text{mean}(\{R_1, R_2, \dots, R_G\})}{\text{std}(\{R_1, R_2, \dots, R_G\})},
\end{equation}
where $\text{mean}(.)$ and $\text{std}(.)$ represent the mean and standard deviation of the rewards, respectively.

\begin{table*}[t]
\small  
\resizebox{\textwidth}{!}{
\begin{tabular}{
p{3.0cm}
>{\centering\arraybackslash}p{1.1cm}
>{\centering\arraybackslash}p{1.1cm}
>{\centering\arraybackslash}p{1cm}
>{\centering\arraybackslash}p{1.1cm}
>{\centering\arraybackslash}p{1.1cm}
>{\centering\arraybackslash}p{1cm}
>{\centering\arraybackslash}p{1.1cm}
>{\centering\arraybackslash}p{1.1cm}
>{\centering\arraybackslash}p{1cm}
}
\toprule
\multirow{2}{*}{\textbf{Model}} & \multicolumn{3}{c}{\textbf{MedChain}} &  \multicolumn{3}{c}{\textbf{Private}} & \multicolumn{3}{c}{\textbf{ClinicalBench}} \\ \cmidrule(lr){2-4} \cmidrule(lr){5-7} \cmidrule(lr){8-10} 
    & \textbf{Precision} & \textbf{Recall} & \textbf{F1}    &\textbf{Precision}  & \textbf{Recall}  & \textbf{F1}     &  \textbf{Precision}   & \textbf{Recall}   & \textbf{F1}  \\ \midrule
\multicolumn{10}{l}{\textit{\textbf{Single-Agent Methods}}}      \\ 
ReAct                  & 26.69 & 23.03 & 21.84 &  17.39 & 17.61 & 14.53 & 24.31 & 11.56& 13.78\\  
ReflecTool            & 28.73 & 38.96 & 29.91 & 21.47 & 29.15 & \underline{21.27} & 31.39 & 28.69 & 26.81 \\  \midrule
\multicolumn{10}{l}{\textit{\textbf{Multi-Agent Methods}}}   \\  
MedAgents              & 18.85& 36.28& 21.82& 18.29& 30.19& 18.69& 22.49& 26.33& 21.22\\
MDAgents               & 27.62& 45.18& 30.30&  19.12& 32.50& 19.79& 32.38& 36.13& 29.77\\
AgentClinic            & 21.13& 21.25& 18.97& 14.68& 16.32& 12.86& 26.00& 14.48& 16.53\\  \midrule

\multicolumn{10}{l}{\textit{\textbf{Agentic Reasoning Methods}}}     \\ 
Tongyi DeepResearch    & 32.15& 35.79& 24.13& 17.92& 30.34& 18.71& 31.10& 38.94& 30.31\\
MedResearcher-R1       & 37.88& 40.09& \underline{30.31}&  21.63& 33.34& 21.19&  37.54 & 33.59 & \underline{31.00}\\ \midrule
\rowcolor[RGB]{236,244,252}
CAREAgent (Ours)       & 26.78 & 51.73 & \textbf{31.75} & 22.92 & 39.81 & \textbf{23.75} & 32.52 & 40.01 & \textbf{31.86}\\ \bottomrule
\end{tabular}
}
\caption{Main Result. The best and second best results are highlighted in \textbf{bold} and \underline{underline}.}
\label{main_result}
\end{table*}

\section{Experiment}


In this section, we first conduct extensive experiments to evaluate the effectiveness of our proposed model. Next, we provide a detailed analysis to offer deeper insights into our proposed model.

\subsection{Training Datasets}

We use the training split of MedChain~\cite{MedChain} as the primary data source, containing 8,514 patient records, and further incorporate a private dataset from the emergency department of a Grade III-A hospital, consisting of 3,000 patient records. 
For patients with multiple courses, we decompose them into multiple single-step instances and summarize the diagnostic and treatment history into the contextual information to preserve essential clinical context. 
Since the original MedChain data is designed for treatment recommendation, we first relabel it as a clinical order generation task using the method described in Appendix~\ref{annotation}. 
We then split the data into training, validation, and test sets with a ratio of 7:1:2.
After a three-stage data filtering process, we finally construct 21,971 high-quality samples with real tool usage for SFT and 3,509 challenging samples for RL training.

\subsection{Evaluation Setup}

\textbf{Benchmarks.}
To comprehensively evaluate model performance, we use the test splits of MedChain and the private dataset, containing 1,050 and 1,098 samples, respectively.
To further assess generalization ability, we additionally evaluate on ClinicalBench, a benchmark designed to evaluate the end-to-end performance of medical agents and LLMs in multi-specialty clinical scenarios. ClinicalBench is not used during training, providing an additional evaluation of model generalization.
Specifically, we select the treatment recommendation task and re-annotate it as a clinical order generation task using the method described in Appendix~\ref{annotation}.

\textbf{Baselines.}
Our baselines fall into three categories: single-agent, multi-agent, and agentic reasoning methods. 
The single-agent baselines include ReAct~\cite{yao2023react} and ReflecTool~\cite{reflectool}, while the multi-agent baselines include MedAgents~\cite{medagents}, MDAgents~\cite{MDAgents}, and AgentClinic~\cite{agentclinic}. 
For the single-agent and multi-agent methods, we adopt Qwen2.5-7B-Instruct~\cite{qwen2.5} as the base model. 
The agentic reasoning baselines include Tongyi DeepResearch~\cite{team2025tongyi} and MedResearcher-R1~\cite{medresearcher}.

%
\textbf{Metrics.}
To evaluate the overall performance of order generation, we use qwen3-next-80b-instruct\footnote{\url{https://qwen.ai/apiplatform}} as the evaluation model to assess clinical order generation, measuring the consistency between predicted orders and reference orders, and computing $Precision$, $Recall$, and $F1$ scores. 
The detailed evaluation prompt is provided in Appendix~\ref{prompt_design}.
Moreover, we employ different evaluation methods for various attributes of the orders. 
For the name, spec, and frequency attributes in both medication and non-medication orders, we use an LLM to assess the accuracy of attribute matching. 
All other attributes are evaluated using \textbf{strict matching} to compute $Accuracy$. 
Detailed experimental results are provided in Appendix A.3.

\begin{figure*}[ht]
    \centering
    \includegraphics[width=1\linewidth]{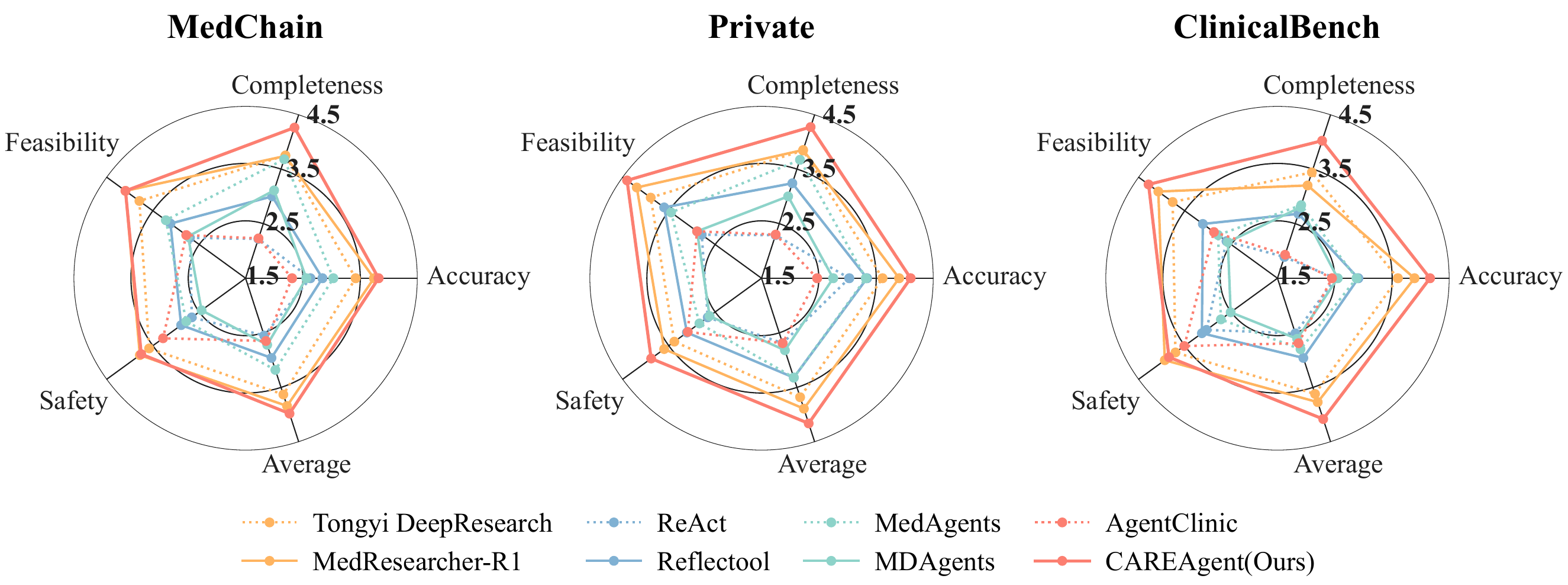}
    \caption{Human Evaluation Results on Completeness, Feasibility, Accuracy, and Safety.}
    \label{radar}
\end{figure*}

\textbf{Implementations.}
We adopt Qwen2.5-7B-Instruct as the backbone model. 
During the SFT stage, we employ the SWIFT framework\footnote{\url{https://github.com/modelscope/ms-swift}} for full-parameter training of the model, with a learning rate of 1e-5, a batch size of 4, and 3 epochs. 
The maximum response length is 8,192 tokens.
In the RL stage, we also use SWIFT, with a learning rate of 5e-5, a warmup ratio of 0.05, and four generations per sample.
The maximum generation length is 8,192 tokens.
The KL regularization coefficient is set to $\beta = 0.05$.
All training experiments are conducted on a cluster of four NVIDIA A800 80GB GPUs.
All experimental results are averaged over three runs.

\subsection{Main Results}




To evaluate the effectiveness of the proposed model, we conduct systematic comparisons with all baseline methods across three benchmarks. 
The experimental results are presented in Table~\ref{main_result}.



From the table, we conclude that:
1) Overall performance across three benchmarks.
Our model achieves the best performance across all three benchmarks. 
Specifically, on MedChain, our method improves the $F1$ score by 1.84\%, 1.45\%, and 1.44\% compared with the strongest Single-Agent, Multi-Agent, and Agentic Reasoning baselines, respectively. 
On ClinicalBench, it achieves further improvements of 5.05\%, 2.09\%, and 0.86\%. 
These gains mainly stem from the proposed agent framework, which constructs high-quality and verifiable reasoning trajectories and leverages them during SFT, thereby improving the model’s ability to capture domain-specific medical knowledge and clinical tool usage patterns. 
In addition, the incorporation of RL further enhances the model’s clinical reasoning capability and cross-dataset generalization.
2) Significant improvements in recall.
Our method also achieves notable improvements in $recall$. 
On MedChain, the $recall$ score increases by 12.77\%, 6.55\%, and 11.64\% compared with the strongest Single-Agent, Multi-Agent, and Agentic Reasoning baselines, respectively. 
Similarly, on ClinicalBench, $recall$ improves by 11.32\%, 3.88\%, and 1.07\%. 
This improvement mainly arises from explicitly modeling the disease treatment decision process in the reasoning trajectory, enabling the model to systematically analyze treatment strategies for each disease and generate a more comprehensive set of medical orders. 

\subsection{Human Evaluation}



In this section, we introduce a human evaluation to further validate the clinical effectiveness and safety of the model’s outputs. Specifically, we compare the proposed framework with the best-performing single-agent method (ReflecTool) and multi-agent method (MDAgents). 
We randomly selected 100 samples and asked a clinician to evaluate them based on a unified set of criteria. The evaluation covers four clinically meaningful dimensions: \textit{Completeness}, \textit{Feasibility}, \textit{Accuracy}, and \textit{Safety}, each with a maximum score of 1. 
\textit{Completeness} assesses whether the orders cover all necessary clinical information; 
\textit{Feasibility} measures whether the orders are executable in real clinical practice;
\textit{Accuracy} checks whether the orders align with current medical knowledge and clinical guidelines;  
\textit{Safety} evaluates the safety of the orders, including attributes such as dosage, route, and potential risks like allergies or priority misclassification.
The experimental results are shown in Fig.~\ref{radar}. From the figure, it is clear that our model significantly outperforms the baseline methods across all four dimensions. 
This advantage stems from the structured, multi-stage reasoning process within our framework, which includes clinical condition assessment, diagnosis, disease stratification, order generation, and prioritization. By defining these stages and basing each decision on comprehensive evidence, our model ensures highly interpretable, logically sound, and clinically relevant recommendations.


\begin{table*}[ht]
\small
\resizebox{\textwidth}{!}{
\begin{tabular}{
p{3.6cm}
>{\centering\arraybackslash}p{1.1cm}
>{\centering\arraybackslash}p{1cm}
>{\centering\arraybackslash}p{0.8cm}
>{\centering\arraybackslash}p{1.1cm}
>{\centering\arraybackslash}p{1cm}
>{\centering\arraybackslash}p{0.8cm}
>{\centering\arraybackslash}p{1.1cm}
>{\centering\arraybackslash}p{1cm}
>{\centering\arraybackslash}p{0.8cm}
}
\toprule
\multirow{2}{*}{\textbf{Model}} & \multicolumn{3}{c}{\textbf{MedChain}}                & \multicolumn{3}{c}{\textbf{Private}}             & \multicolumn{3}{c}{\textbf{ClinicalBench}}           \\ \cmidrule(lr){2-4} \cmidrule(lr){5-7} \cmidrule(lr){8-10} & \textbf{Precision} & \textbf{Recall} & \textbf{F1} & \textbf{Precision} & \textbf{Recall} & \textbf{F1} & \textbf{Precision} & \textbf{Recall} & \textbf{F1} \\ \midrule

\multicolumn{10}{l}{\textit{\textbf{Remove Different Module of Agent Framework}}}  \\
\emph{w/o} Tool   & 27.30& 30.03& 24.72&      22.60& 24.56& 19.11&     21.53& 12.94& 14.09\\
\emph{w/o} 4-stage reasoning    & 29.60& 28.16& 25.57&  21.81& 20.91& 17.92&   28.95& 15.27& 17.49\\
\emph{w/o} Tool \& 4-stage reasoning   & 31.41& 26.33& 24.86&   28.23& 24.37& 21.66&   31.86& 13.85& 16.76\\ \midrule

\multicolumn{10}{l}{\textit{\textbf{Training with Template-Based Synthetic Data}}}  \\
\emph{w/o} SFT   & 24.05& 33.01& 24.49&      15.82& 22.66& 15.48&     24.76& 22.24& 19.69\\
\emph{w/o} RL    & 9.27& 27.64& 12.83&  11.89& 31.99& 14.02&   18.97& 36.48& 22.51\\
SFT+RL           & 9.32& 28.87& 12.98&   10.98& 31.61& 13.35&   19.24& 35.25& 22.35\\ \midrule

\multicolumn{10}{l}{\textit{\textbf{Training with Full Data}}}     \\
\emph{w/o} SFT   & 29.61& 35.46& 28.38&      22.33& 25.93& 19.96&     25.16& 17.45& 18.08\\
\emph{w/o} RL    & 26.08& 48.14& \underline{30.72}&    23.64& 39.82& \underline{23.67}&   28.69& 34.83& 27.97\\
SFT+RL           & 25.53& 48.16& 30.16&     21.01& 35.46& 21.71&   30.27& 36.89& \underline{29.58} \\ \midrule
\multicolumn{10}{l}{\textit{\textbf{Training with Agent-Based Synthetic Data}}}   \\
Qwen2.5-7B-Instruct    & 29.43& 35.44& 28.22&      21.60& 25.80& 19.56&     24.79& 17.05& 17.68\\
\emph{w/o} SFT         & 25.27& 33.56& 25.83&      16.51& 23.21& 16.22&     24.36& 19.75& 18.64\\
\emph{w/o} RL          & 25.63& 46.78& 29.99&      22.47& 38.58& 23.24&     28.87& 35.46& 28.29\\
SFT+RL            & 26.78 & 51.73 & \textbf{31.75} & 22.92 & 39.81 & \textbf{23.75} & 32.52 & 40.01 & \textbf{31.86}\\ \bottomrule

\end{tabular}
}
\caption{Ablation study on agent framework, agentic reasoning data construction, and post-training strategies.}
\label{ablation_study}
\vspace{-0.2cm}
\end{table*}

\subsection{Ablation Study}

To validate the effectiveness of the proposed agent framework, data construction method, and post-training strategy, we conduct a series of ablation studies. 
For the agent framework, we conduct ablation studies by removing the tool-use module, the four-stage reasoning process, and both components simultaneously, to assess their contributions to overall performance.
For the agentic reasoning data construction method, we further design two variants: (1) template-based synthetic trajectories and (2) a hybrid approach that combines template-based synthetic trajectories with agent-based synthesized trajectories (denoted as Full Data).
In addition, we separately ablate SFT and RL to assess the contribution of each post-training component.
The experimental results are reported in Table~\ref{ablation_study}.


From the table, we conclude that:
1) Sequentially removing components from the agent framework consistently reduces performance, demonstrating the effectiveness of both the tool-use module and the four-stage reasoning process. 
On MedChain, $F1$ drops from 31.75 to 24.72 without the tool and to 25.57 without the reasoning stages, further declining to 24.86 when both are removed. 
This is because the tool integrates external clinical knowledge, while four-stage reasoning enables systematic patient assessment, diagnosis, order generation, and prioritization, supporting precise and appropriate decisions.
2) Agent-based synthetic trajectories outperform template-based ones.
On MedChain, training with SFT and RL on agent-based data improves $F1$ by 18.77\% and 1.59\% over models trained on template-based and full data, respectively. This advantage arises because agent-based synthesis captures authentic, complete reasoning trajectories, whereas template-based methods are constrained by predefined patterns.
3) Removing components from the post-training pipeline leads to performance degradation. 
On MedChain, directly applying RL to Qwen2.5-7B decreases $F1$ by 2.39\%, while incorporating SFT yields a 1.77\% gain; combining SFT with RL further increases $F1$ by 3.53\%.  
This reflects that the base model’s limited medical knowledge and reasoning understanding makes direct RL unstable, whereas SFT provides foundational knowledge and format understanding, enabling RL to effectively enhance reasoning capability.

\section{Conclusion}


This paper introduces CAREAgent, which enables LLMs to autonomously invoke multiple external tools during step-by-step reasoning. 
To construct training data, we design a two-stage agentic data construction pipeline that first collects real reasoning trajectories within an agent framework and then applies filtering and selection to obtain high-quality data for SFT and RL.
During training, the model is first fine-tuned via SFT to learn structured reasoning and tool usage patterns, and is subsequently optimized with RL to further enhance its clinical reasoning capabilities. Extensive experiments across three benchmarks demonstrate the effectiveness and robustness of CAREAgent.

\section*{Limitations and Future Work}


Although the proposed method demonstrates strong effectiveness, several limitations remain and warrant further investigation.
First, the overall performance of CAREAgent still falls short of that of experienced clinical practitioners. Therefore, it is better suited as a clinical decision-support tool rather than a fully autonomous clinical decision-making system.
Second, CAREAgent currently integrates only two external tools, Search and Clinical\_Assessment. 
Given the rapidly expanding tool ecosystem, incorporating additional tools, such as vision–language models, could further enhance medical image understanding and multimodal clinical reasoning.
In future work, we plan to continue expanding the range of integrated tools to advance multi-tool collaborative reasoning.
Finally, due to limited computational resources and the high cost of multi-tool reasoning, this work primarily focuses on 7B-parameter models. In future research, we aim to extend our experiments to larger models (e.g., 32B) to evaluate performance on more complex tasks and assess the generalizability of our approach across model scales.

\section*{Ethics Consideration}

MedChain and ClinicalBench are publicly available datasets and do not involve any private or sensitive personal information. For the private data used in this study, all EHRs and associated codes were processed in strict accordance with relevant guidelines, undergoing double de-identification by both an ethics committee and domain experts, and have been approved by the institutional review board of our partner hospital.
During data processing, we implemented rigorous anonymization procedures to ensure that no protected health information or identifiable patient data is included. As all datasets used are either publicly available or fully de-identified, and contain no personally identifiable information, we further conducted careful assessments of the potential impacts and risks of this work to prevent misuse or unintended harm. Therefore, this study fully complies with applicable ethical standards.



\bibliography{custom}

\appendix

\clearpage

\section{Appendix}
\label{sec:appendix}

\subsection{Clinical Order Annotation Process}
\label{annotation}

In this section, we detail the annotation process for clinical orders.
For medication orders, we use DeepSeek-V3.1~\cite{liu2024deepseek} to extract key attributes from the raw medical records, including \textit{order names}, \textit{duration days}, \textit{frequency}, \textit{drug dosage}, and \textit{route of administration}. 
For non-medication orders, we extract attributes such as \textit{order name}, \textit{order type}, \textit{duration type}, and \textit{frequency}. 
To mitigate hallucinations, we explicitly instruct the model to retain only information explicitly stated in the medical records and strictly prohibit any fabricated content.
For the \textit{priority} of order, we categorize the priority levels into six grades, from P1 to P6, and have developed a two-step annotation process. 
Specifically, the first step involves hard constraints, where the highest priority (P1) is assigned to interventions for critical conditions (such as ventilation, vasopressors, hemostasis, and correction of life-threatening disturbances). The second step involves soft constraints, where we use the DeepSeek-V3.1 model to score each clinical order based on four dimensions: Urgency, Impact, SafetyBuffer, and EffortFactor. The specific scoring criteria are as follows:
\begin{itemize}
    \item \textbf{Urgency:} Measures the time sensitivity and clinical urgency of the clinical order, with a score range of 0-1.
    \item \textbf{Impact:} Assesses the actual impact of the order on the patient’s condition, prognosis, and treatment path, with a score range of 0-1.
    \item \textbf{SafetyBuffer:} Represents the allowable time deviation or delay when executing the order. The smaller the safety buffer (indicating higher risk), the higher the score, with a score range of 0-1.
    \item \textbf{EffortFactor:} Represents the cost or effort required to execute the order, with a score range of 0-1.
\end{itemize}
Subsequently, we assign different weights to each score and calculate the total priority score:
\begin{equation}
\begin{aligned}
    Priority = 0.45 \times \text{Urgency} + 0.35 \times \text{Impact} \\
    + 0.15 \times \text{SafetyBuffer} + 0.05 \times \text{EffortFactor}.
\end{aligned}
\end{equation}
Based on this score, clinical orders are assigned to priority levels ranging from P1 to P6. 
An overview of each priority level is provided in Table~\ref{priority_definition}.
The priority score takes into account factors such as urgency, impact, safety buffer, and execution cost, enabling a more nuanced adjustment of priorities.
For the \textit{rationale} of order, clinical doctors initially constructed three seed templates. 
We then used the DeepSeek-V3.1 model to generate the corresponding Reason for each clinical order. Finally, 20\% of the annotated data is randomly sampled for quality checks, and any samples with accuracy below 95\% are manually corrected by the clinical doctors.

To assess the quality of our data annotation, we conducted a clinician consistency evaluation. 
We randomly sampled 100 cases and asked experienced clinicians to independently annotate the data.
The results show a high level of agreement with our constructed labels, with Cohen’s kappa scores of 0.64 for MedChain and 0.71 for ClinicalBench.

\begin{table}[!t]
\centering
\small
\resizebox{\linewidth}{!}{
\begin{tabular}{p{0.5cm} p{2.2cm} p{3cm} p{1.8cm}}
\toprule
\textbf{Level} & \textbf{Name} & \textbf{Clinical Meaning} & \textbf{Response Time} \\
\midrule
P1 & Critical or STAT  &
Immediate life-saving interventions for acute instability or imminent threat to life &
Immediate ($\leq$ 5 min) \\ \midrule

P2 & Emergent or Urgent Critical Support &
Core treatments required to maintain vital functions or prevent rapid deterioration &
$\leq$ 30 min \\ \midrule

P3 & Important Monitoring and Follow-up &
Dynamic assessment of treatment efficacy or monitoring of key risk indicators &
Execute within $\leq$ 2 h (dependent) \\ \midrule

P4 & Routine Therapeutic and Maintenance Care &
Scheduled execution of routine or maintenance therapeutic interventions &
Same day ($\leq$ 6 h) \\ \midrule

P5 & Nursing and Prophylactic Care &
Nursing operations or preventive interventions &
$\leq$ 8 h \\ \midrule

P6 & Planned or Non-Urgent Care &
Deferrable interventions executed according to planned conditions &
Plan or execute within $\leq$ 24 h \\
\bottomrule
\end{tabular}
}
\caption{Definitions of Clinical Order Priority Levels.}
\label{priority_definition}
\end{table}

\subsection{Analysis of Different Priority Levels}

\begin{figure}[!t]
    \centering
    \includegraphics[width=\linewidth]{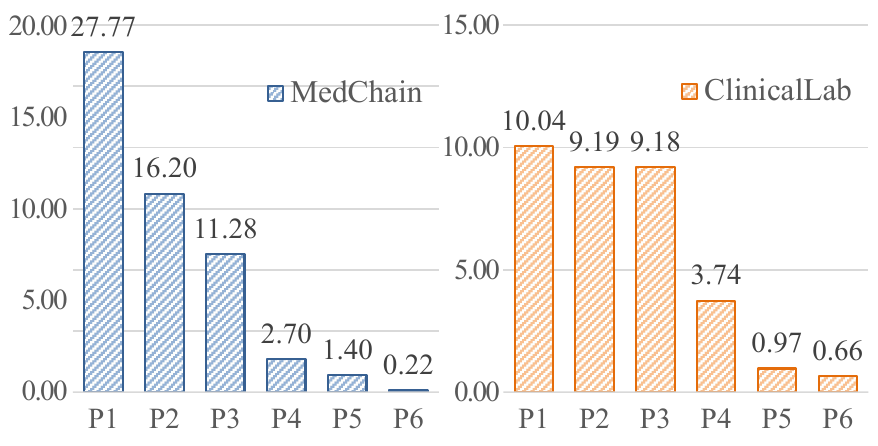}
    \caption{F1 Score by Different Order Priority Level.}
    \label{placeholder}
\end{figure}

To analyze the model’s performance across different clinical order priorities, we group orders by priority and evaluate overall order generation performance for each level from P1 to P6. 
The experimental results are shown in Figure~\ref{placeholder}. 
As observed, the model performs substantially better on high-priority orders (P1–P3) than on lower-priority ones (P4–P6). 
This result indicates that, in clinical scenarios involving patient safety and urgent interventions, the model demonstrates a strong ability to identify critical information and make more appropriate high-priority decisions.

\subsection{Analysis of Clinical Order Attributes}
\label{analysis_attribute}



To further analyze the model's performance across different clinical order attributes, we select the best-performing baseline for each method in the main experiment as reference. For both medication and non-medication orders, the order name, drug dosage, and frequency attributes are evaluated via LLM-based semantic matching, while the remaining attributes are evaluated via strict exact matching. The results are reported in Table~\ref{medchain_attributes} and Table~\ref{clinicalbench_attributes}.
The experimental results show that our model achieves notable improvements on several key attributes for both medication and non-medication orders, particularly Order Name, Frequency, and Drug Dosage. For medication orders, Order Name demonstrates the largest gain, increasing from 43.95 to 51.37 on MedChain and from 47.71 to 55.97 on ClinicalBench. For non-medication orders, Frequency exhibits the most substantial improvement, rising from 8.52 to 18.02 on MedChain and from 13.96 to 18.0 on ClinicalBench. Other attributes, including Duration and Route, generally benefit from our method, although some attributes such as Order Priority and certain Route/Drug Dosage entries may still be outperformed by specific baselines in certain datasets.
These improvements primarily stem from two factors. First, the four-stage reasoning process enables the model to analyze each attribute at finer granularity, enhancing core content accuracy. Second, retrieval-augmented generation provides rich background knowledge and similar-case information, improving understanding of specific attributes and enabling more precise output generation.

\begin{table}[ht]
\centering
\renewcommand{\arraystretch}{1} 
\setlength{\tabcolsep}{0pt}
\resizebox{\linewidth}{!}{
\begin{tabular}{%
    l
    m{1.8cm}<{\centering}
    m{2.2cm}<{\centering}
    m{3cm}<{\centering}
    m{1.2cm}<{\centering}
}
\toprule
\textbf{Attributes} & \textbf{Reflectool} & \textbf{MDAgents} & \textbf{\makebox[2cm]{MedResearcher-R1}} & \textbf{Ours} \\
\midrule
\multicolumn{5}{l}{\textit{\textbf{Medication Order}}} \\
Order Name & 38.78 & 43.95 & 34.24 & 51.37 \\
Duration Days & 4.54 & 0.37 & 4.27 & 12.12 \\
Frequency & 9.99 & 9.33 & 9.71 & 16.64 \\
Drug Dosage & 8.23 & 7.95 & 7.25 & 12.49 \\
Route of Administration & 17.2 & 16.64 & 11.57 & 24.02 \\
Order Priority & 16.88 & 19.52 & 11.95 & 15.24 \\
\midrule
\multicolumn{5}{l}{\textit{\textbf{Non-medication Order}}} \\
Order Name & 45.33 & 60.16 & 49.23 & 59.47 \\
Order Type & 12.89 & 16.53 & 16.68 & 24.2 \\
Duration Type & 11.19 & 16.07 & 16.74 & 23.38 \\
Frequency & 4.52 & 8.52 & 10.88 & 18.02 \\
Order Priority & 10.83 & 13.91 & 13.86 & 17.56 \\
\bottomrule
\end{tabular}
}
\caption{Accuracy of Clinical Order Attributes on MedChain.}
\label{medchain_attributes}
\end{table}

\begin{table}[ht]
\centering
\renewcommand{\arraystretch}{1} 
\setlength{\tabcolsep}{0pt}
\resizebox{\linewidth}{!}{
\begin{tabular}{%
    l
    m{1.8cm}<{\centering}
    m{2.2cm}<{\centering}
    m{3cm}<{\centering}
    m{1.2cm}<{\centering}
}
\toprule
\textbf{Attributes} & \textbf{Reflectool} & \textbf{MDAgents} & \textbf{\makebox[2cm]{MedResearcher-R1}} & \textbf{Ours} \\
\midrule
\multicolumn{5}{l}{\textit{\textbf{Medication Order}}} \\
Order Name & 38.4 & 47.71 & 31.17 & 55.97 \\
Duration Days & 4.79 & 0.05 & 9.82 & 10.63 \\
Frequency & 17.05 & 18 & 12.9 & 21.84 \\
Drug Dosage & 17.68 & 22.1 & 15.78 & 17.3 \\
Route of Administration & 24.62 & 30.39 & 20.03 & 28.29 \\
Order Priority & 12.99 & 16.29 & 8.74 & 9.76 \\
\midrule
\multicolumn{5}{l}{\textit{\textbf{Non-medication Order}}} \\
Order Name & 24.39 & 38.23 & 23.71 & 39.32 \\
Order Type & 10.03 & 18.33 & 13.61 & 18.53 \\
Duration Type & 9.7 & 21.47 & 12.22 & 21.22 \\
Frequency & 9.01 & 13.96 & 12.3 & 18 \\
Order Priority & 5.82 & 6.91 & 4.87 & 5.72 \\
\bottomrule
\end{tabular}
}
\caption{Accuracy of Clinical Order Attributes on ClinicalBench.}
\label{clinicalbench_attributes}
\end{table}

\subsection{Analysis of Clinical Order Types}
To evaluate model performance across clinical order types, we categorize orders into \emph{medication} and \emph{non-medication} orders, with the latter further divided into laboratory tests, examinations, treatments, operations, nursing care, and nutrition. 
The results are summarized in Table~\ref{order_type}. 
As shown in the table, performance varies across order types. 
The model achieves relatively high precision but low recall on medication orders.
For non-medication orders, it performs more consistently and achieves better overall results on laboratory test, examination, and operation orders, while performance on nursing care, nutrition, and other orders is comparatively weaker.


\begin{table}[!t]
\resizebox{\linewidth}{!}{
\begin{tabular}{lcccccc}
\toprule
\multirow{2}{*}{\textbf{Type}} & \multicolumn{3}{c}{\textbf{MedChain}}              & \multicolumn{3}{c}{\textbf{ClinicalBench}}         \\ \cmidrule(lr){2-4}  \cmidrule{5-7}
& \textbf{P} & \textbf{R} & \textbf{F1} & \textbf{P} & \textbf{R} & \textbf{F1} \\ \midrule
\multicolumn{7}{l}{\textit{\textbf{Medication Order}}}  \\
Medication                     & 75.62& 41.31& 39.48& 64.86& 40.65& 37.28\\ \midrule
\multicolumn{7}{l}{\textit{\textbf{Non-medication Order}}}  \\
Laboratory Test                & 65.51& 37.98& 31.74& 87.96& 27.41& 28.99\\
Examination                    & 72.27& 25.92& 25.07& 75.21& 15.08& 15.85\\
Treatment                      & 68.29& 23.39& 22.07& 51.03& 14.61& 13.89\\
Operation             & 82.65& 45.16& 41.67& 93.81& 55.52& 54.06\\
Nursing Care                   & 23.26& 22.50& 13.15& 41.92& 21.97& 17.21\\
Nutrition                      & 25.00& 14.29& 10.00& 34.78& 11.94& 10.00\\ \bottomrule
\end{tabular}
}
\caption{Results for Each Order Type. ``P'' and ``R'' denote ``Precision'' and ``Recall''.}
\label{order_type}
\end{table}

\begin{table}[!t]
\centering
\resizebox{\linewidth}{!}{
\begin{tabular}{lcccccc}
\toprule
\multirow{2}{*}{\textbf{Frequency}} 
& \multicolumn{3}{c}{\textbf{MedChain}} 
& \multicolumn{3}{c}{\textbf{ClinicalBench}} \\
\cmidrule(lr){2-4} \cmidrule(lr){5-7}
& \textbf{P} & \textbf{R} & \textbf{F1}
& \textbf{P} & \textbf{R} & \textbf{F1} \\
\midrule
0--1 & 23.80& 42.20& 28.17& 29.42& 30.32& 26.84\\
2--3 & 26.00& 47.48& 30.38& 28.80& 35.63& 28.30\\
4--9 & 3.97& 9.01& 5.29& 9.08& 10.33& 8.86\\
\bottomrule
\end{tabular}
}
\caption{Results by Tool Invocation Frequency.}
\label{tool_frequency}
\end{table}

\subsection{Analysis of Tool Invocation Frequency}
To analyze the impact of tool invocation frequency within trajectories on model performance, we evaluate the model under different numbers of tool calls. Based on statistics collected from the MedChain and ClinicalBench, the maximum number of tool invocations per trajectory is 9. We therefore group tool invocation frequency into three ranges: 0–1, 2–3, and 4–9 times. The experimental results are presented in Table~\ref{tool_frequency}.
From the table, we observe that tool invocation frequency has a clear impact on model performance. In both MedChain and ClinicalBench, trajectories containing 2–3 tool calls achieve the best overall performance, indicating that moderate tool usage helps the agent obtain sufficient external evidence while maintaining coherent reasoning. In contrast, trajectories with 0–1 tool calls show slightly lower performance, suggesting that insufficient tool interaction may limit the model’s access to necessary clinical knowledge. Meanwhile, when the number of tool invocations increases to 4–9 times, performance drops significantly across all metrics. This indicates that excessive tool usage may introduce redundant or noisy information and disrupt the reasoning process. Overall, these results suggest that a moderate number of tool interactions is most beneficial for clinical decision generation, while overly frequent tool calls may negatively affect reasoning stability.


\begin{table}[!t]
\centering
\small
\resizebox{\linewidth}{!}{
\begin{tabular}{
p{1.8cm}
>{\centering\arraybackslash}p{1.7cm}
>{\centering\arraybackslash}p{1.5cm}
>{\centering\arraybackslash}p{1.2cm}
}
\toprule
\textbf{\# Courses} & \textbf{Precision} & \textbf{Recall} & \textbf{F1} \\
\midrule
1--3    & 24.55& 46.66& 29.28\\
4--6    & 27.71& 51.53& 31.81\\
$>$7    & 25.74& 42.01& 28.82\\
\bottomrule
\end{tabular}
}
\caption{Performance under Different Numbers of Courses.}
\label{episode_num}
\end{table}

\subsection{Analysis of Multiple Courses}


To analyze the impact of the number of courses on model performance, we evaluate the model across different courses counts. 
As most samples in the ClinicalBench dataset contain only a single course, our analysis primarily focuses on the MedChain dataset. We partition samples into three groups based on the number of courses: 1–3, 4–6, and more than 7 courses. The experimental results are presented in Table~\ref{episode_num}.
As shown in the table, the number of courses has a clear impact on model performance. 
The model achieves the best results when the number of courses is 4–6, obtaining the highest Precision, Recall, and F1 scores. 
This indicates that the framework benefits from a moderate amount of longitudinal clinical information, where multiple courses provide sufficient evidence for understanding disease progression and generating appropriate medical orders.
When the number of courses is 1–3, performance is slightly lower, which is likely due to the limited temporal information available for clinical reasoning. 
In contrast, when the number of courses exceeds 7, the performance shows a moderate decline. 
This is expected because longer clinical trajectories typically involve more complex patient conditions and introduce additional reasoning difficulty. 
Notably, the performance degradation remains relatively limited, suggesting that the case summarization agent effectively compresses newly added course information and mitigates context growth, allowing the framework to maintain stable reasoning even under long clinical histories.

\begin{table}[!t]
\centering
\small
\resizebox{\linewidth}{!}{
\begin{tabular}{
p{1.2cm}
>{\centering\arraybackslash}p{1.6cm}
>{\centering\arraybackslash}p{1.2cm}
>{\centering\arraybackslash}p{1.6cm}
>{\centering\arraybackslash}p{0.6cm}
}
\toprule
\textbf{Dataset} & \textbf{Total Samples} & \textbf{Avg. Tokens} & \textbf{Inference Time (min)} & \textbf{F1} \\
\midrule
ClinicalBench & 1500 & 2056.68 & 37 & 31.86 \\
MedChain     & 1050 & 1647.70 & 26 & 31.75 \\
\bottomrule
\end{tabular}
}
\caption{Inference efficiency and performance across different datasets.}
\label{dataset_inference}
\end{table}

\subsection{Analysis of Efficiency}

To analyze the efficiency of our framework, we evaluate inference time and performance on two datasets with different sample sizes and input lengths. The results are shown in Table~\ref{dataset_inference}.
As shown in the table, the framework processes the ClinicalBench and MedChain datasets in 37 and 26 minutes, respectively, covering 2550 samples in total. Despite differences in average token length (1647–2056 tokens), the inference time scales proportionally with the dataset size, indicating stable efficiency across datasets with varying input lengths.
In terms of performance, the model achieves comparable F1 scores on both datasets, demonstrating that the framework maintains consistent effectiveness while handling relatively long clinical contexts. Overall, these results show that the proposed method provides stable inference efficiency and good scalability for real-world clinical decision support tasks.

\subsection{Analysis of Reward Weights in RL}



To investigate the influence of reward weights on Agentic RL performance, we performed a weight sensitivity analysis. The reward function is divided into three components: Format, Reasoning, and Answer. In addition to the original weights (1:3:6), we considered two alternatives: Variant 1 (1:1:1) and Variant 2 (1:6:3). The results are shown in Table~\ref{reward_weights}.
The results show that Ours consistently outperforms both variants across all three datasets. On MedChain, Private, and ClinicalBench, the F1 scores of Ours are 31.75, 23.75, and 31.86, representing improvements of 1.39, 1.01, and 4.00 over Variant 1, and 2.18, 0.82, and 4.17 over Variant 2, respectively.
Variant 1 distributes reward weights uniformly during RL, yet its effect is limited. This suggests that, once format and basic reasoning are mastered in the SFT stage, RL should focus on complex reasoning and answer optimization. Variant 2 assigns more weight to the reasoning component but achieves suboptimal performance, indicating that for tasks where answer accuracy is the final metric, overemphasis on reasoning rewards can generate superficially coherent reasoning chains that undermine actual answer quality. In contrast, our weight configuration effectively improves answer accuracy while preserving reasoning quality, achieving the highest performance across all datasets.

\begin{table}[!t]
\centering
\setlength{\tabcolsep}{2.5pt}
\resizebox{\linewidth}{!}{
\begin{tabular}{lccccccccc}
\toprule
\multirow{2}{*}{\textbf{Weights}} 
& \multicolumn{3}{c}{\textbf{MedChain}} 
& \multicolumn{3}{c}{\textbf{Private}} 
& \multicolumn{3}{c}{\textbf{ClinicalBench}} \\
\cmidrule(lr){2-4} \cmidrule(lr){5-7} \cmidrule(lr){8-10}
& \textbf{P} & \textbf{R} & \textbf{F1} 
& \textbf{P} & \textbf{R} & \textbf{F1} 
& \textbf{P} & \textbf{R} & \textbf{F1} \\
\midrule
Variant 1 & 25.67 & 47.99 & 30.36 & 22.37 & 35.54 & 22.74 & 28.76 & 34.60 & 27.86 \\
Variant 2 & 24.65 & 48.17 & 29.57 & 22.56 & 37.00 & 22.93 & 28.34 & 34.38 & 27.69 \\
Ours & 26.78 & 51.73 & 31.75 & 22.92 & 39.81 & 23.75 & 32.52 & 40.01 & 31.86 \\
\bottomrule
\end{tabular}
}
\caption{Training Performance of Agentic RL under Different Reward Weights.}
\label{reward_weights}
\end{table}

\subsection{Workflow of the Summarization Agent}
\label{summary}

In this section, we provide a detailed explanation of the workflow of the summary agent. We have developed a two-stage approach for summarization. 
The first stage is patient condition categorization, where the goal is to classify the patient’s daily medical records and test results into predefined condition categories. 
The second stage is treatment summary generation, where we aim to select the appropriate summary template based on the condition category and generate a summary of the patient’s daily medical records and test results. 
Through statistical analysis of the patient’s daily clinical records, we identified six common condition categories and developed corresponding summary templates for each category. 
The detailed categories and templates are shown in Table~\ref{summary_category}.

\begin{table*}[t]
\small

\resizebox{\textwidth}{!}{
\begin{tabular}{p{3cm}p{4.5cm}p{7.3cm}}
\toprule
\textbf{Category}  & \textbf{Category Description}   & \textbf{Summary Template} \\ \midrule
Sudden Deterioration / Acute Exacerbation of the Patient’s Condition                 & This category is primarily used to document acute changes in a patient’s condition or sudden clinical events (such as respiratory distress, altered consciousness, or abrupt changes in vital signs), as well as the emergency interventions undertaken in response to these abnormalities.                      &  When a patient presents with abnormal clinical signs (e.g., sudden dyspnea, altered consciousness, or decreased oxygen saturation), physical examination and relevant tests (such as blood gas analysis, imaging, or lab results) reveal corresponding abnormalities. A comprehensive assessment suggests these changes may be linked to underlying causes or pathophysiological mechanisms (e.g., infection progression or respiratory failure). In response, emergency interventions are promptly initiated, including airway suctioning, oxygen therapy, endotracheal intubation, mechanical ventilation, and enhanced anti-infective or supportive treatment. Following these measures, the patient’s condition may improve, stabilize, or remain under observation. \\ \midrule
The patient’s condition remained stable / improved / recovered on that day.          & This category is used to describe improvement or recovery in the patient’s symptoms, physical findings, or examination results following treatment, with no significant new abnormalities identified. It also documents any subsequent adjustments to the treatment plan.         & After receiving treatments (e.g., anti-infective therapy, respiratory support, and nutritional support), the patient’s symptoms have improved, signs are stable, and laboratory markers show progress. As the patient’s condition stabilizes or improves, the original treatment plan will continue, with gradual reduction or discontinuation of specific medications or interventions as needed. The patient will be closely monitored, and measures to prevent complications will remain in place.   \\ \midrule

Important Examinations / Results Update                                              & This category is used to highlight updates on newly available key laboratory, imaging, or other diagnostic test results, and to document any corresponding adjustments to the diagnosis or treatment plan.      & 
To assess the patient’s condition and treatment effectiveness, today’s examinations (e.g., chest X-ray, CT scan, blood gas analysis, biochemistry, complete blood count) revealed key findings (e.g., worsening bilateral lung effusion, elevated white blood cell count, reduced pleural effusion, improved cardiac function). Based on these results, the impact on the diagnosis and treatment plan was evaluated (e.g., infection control, antibiotic adjustments, need for enhanced support). As a result, decisions were made to adjust, continue, or discontinue specific treatments, and a follow-up plan or re-examination schedule was set. \\ \midrule

Department Transfer / Level of Care Change / Consultation / Nursing-Related Care     & This category is used to document instances where a patient is transferred to another department or monitoring unit due to their condition or treatment needs, or when consultations with relevant specialists or enhanced nursing measures are required.               &       Given the patient’s condition and treatment needs (e.g., advanced monitoring, specialist evaluation, or nursing interventions), a decision was made to transfer the patient to the appropriate department or ICU, request specialist consultations, or enhance nursing care. This will facilitate better monitoring, further treatment, and additional diagnostic tests. Recommendations, such as enhanced suctioning, pain management, psychological support, and special positioning, have been implemented and are continuously assessed for effectiveness.     \\ \midrule

New Diagnosis Established / Diagnosis Updated                                        & This category is used to establish new diagnoses, revise previous diagnoses, or identify comorbidities, and to adjust the treatment strategy accordingly.             &      Based on the latest clinical presentation, laboratory results, imaging data, and specialist consultations, the patient has been diagnosed with a new or revised diagnosis (e.g., secondary infection, atelectasis, psychiatric disorder, etc.). This diagnosis suggests possible underlying pathological mechanisms or prognostic implications, prompting adjustments to the treatment plan, including the addition or modification of medications or interventions (e.g., antifungal therapy, anticoagulation, psychiatric medications, etc.). A follow-up plan and further examinations have been scheduled to assess treatment efficacy and monitor changes in the patient’s condition.   \\ \midrule
Other Matters (e.g., special situations, family communication, and informed consent) & This category is used for documenting important situations that involve communication with family members, informed consent, special events (such as critical illness, refusal or discontinuation of treatment, etc.), or other significant circumstances that do not fall under the categories mentioned above.      &   Throughout diagnosis and treatment, given the patient’s special circumstances (e.g., critical illness, treatment decisions, complications, financial concerns), timely communication with the family was made to explain the patient’s condition, treatment risks, and expected outcomes. With the family’s consent, specific measures were agreed upon (e.g., consent for intubation, discontinuation of treatments, signing the informed consent form), and all necessary procedures were completed. The patient’s condition and the family’s wishes will continue to be closely monitored.   \\ \bottomrule
\end{tabular}
}

\caption{
Six Types of Medical Condition Categories and Their Corresponding Summary Templates.
}
\label{summary_category}
\end{table*}

\subsection{Details of Tool Design}


To provide a more detailed description of the tool design and usage guidelines, we present the implementation details and usage specifications of the Search and Clinical Scoring tools used in this work.

\begin{itemize}
    \item \textbf{Search Tools:} Provide vector-based retrieval over clinical guidelines, medical literature corpora, and authoritative medical textbooks, enabling efficient and accurate medical knowledge acquisition for clinical decision support. Specifically, we collect a total of 250 documents as the knowledge base and deploy them using the RAGFlow\footnote{\url{https://github.com/infiniflow/ragflow}} framework. The knowledge base consists of 250 documents, including 186 in Chinese and 64 in other languages, with file types comprising 113 PDFs, 99 DOCX files, and 38 DOC files. These documents cover the period from 2009 to 2023.
    \item \textbf{Clinical Scoring Tools:} Support the computation of a wide range of clinical scale scores, risk assessment metrics, and disease severity indices through dedicated computational modules or standalone clinical calculators. Specifically, these tools include mean arterial pressure (MAP), HEART score, Wells score for pulmonary embolism, PERC rule for pulmonary embolism, CURB-65, QTc correction (Bazett and Fridericia), CKD-EPI (or Cockcroft–Gault), anion gap (with albumin correction), EDACS score, Geneva score, PESI score, YEARS score, and the aortic dissection risk score (ADRS).
\end{itemize}


\subsection{Case Study}



To clearly illustrate how CAREAgent performs clinical reasoning through tool invocation, we conduct a case study that explicitly demonstrates its reasoning process. 
The model’s thought process is enclosed within the \purple{<thought>} and \purple{</thought>} tags; tool calls and their parameters are presented within \green{<tool\_call>} and \green{</tool\_call>}; the results returned by the tools are included within \blue{<observation>} and \blue{</observation>}; and the final generated output is provided within \red{<answer>} and \red{</answer>}.
The clinical records and corresponding reference orders from ClinicalBench are presented in Table~\ref{case_clinicalbench}, and the reasoning trajectory of CAREAgent on this example is shown in Table~\ref{CAREAgent_clinicalbench}. 
The clinical records and reference orders from MedChain are provided in Table~\ref{case_medchain}. 
As this dataset contains multiple courses for each patient, we select a single course as an illustrative example; accordingly, the record includes a summary of the patient’s prior clinical history. 
The reasoning trajectory of CAREAgent for this example is shown in Table~\ref{CAREAgent_medchain}.


\begin{table*}[!t]
    \centering
    
    \fontsize{9pt}{11pt}\selectfont
    \begin{tabular}{p{1\linewidth}}
    \toprule
        \rowcolor{gray!20}\textbf{Example \#1 from ClinicalBench Dataset} \\
    \midrule
        \textbf{EMR:} \\ ``clinical\_department'': ``Department of Cardiology'',\\
          ``principal\_diagnosis'': ``Acute Myocardial Infarction'',\\
          ``preliminary\_diagnosis'': ``1. Comminuted intertrochanteric fracture of the left femur (four-part fracture); 2. Grade 3 hypertension (high risk); 3. Acute myocardial infarction; 4. Severe osteoporosis.'',\\
          ``diagnostic\_basis'': ``1. The patient presented with left hip pain and limited mobility for 3 hours following a fall. 2. Physical examination findings support the diagnosis: physiological spinal curvature was preserved with no tenderness over the spinous processes; assessment of spinal mobility in the supine position could not be performed due to poor cooperation. Mild swelling was observed in the left hip, with external rotation and shortening deformity of the left lower limb (approximately 80° of external rotation). There was tenderness around the left hip, and the patient refused palpation and movement due to pain. Movement of the left knee, ankle, and toes was preserved. Pulses of the left dorsalis pedis and posterior tibial arteries were palpable. No obvious abnormalities were observed in both upper limbs or the contralateral lower limb. 3. Imaging findings support the diagnosis: echocardiography revealed segmental wall motion abnormalities, left ventricular enlargement, mild interventricular septal thickening, slight dilation of the main pulmonary artery, globally reduced left ventricular function, moderate mitral regurgitation, mild aortic regurgitation, and a small amount of pericardial effusion. 4. Laboratory findings support the diagnosis: complete blood count showed decreased red blood cell count, hemoglobin, and hematocrit, increased red cell distribution width, and elevated C-reactive protein.'',\\
          ``differential\_diagnosis'': ``The diagnosis is definitive, and no differential diagnosis is required.'' \\ \\
        \textbf{Labeled Answer:} \\
        \{ \\
        ``name'': ``Complete Blood Count'', ``type'': ``Laboratory Test'',
            ``duration\_type'': 2, ``frequency'': ``None'',
            ``reason'': ``Given the patient’s acute myocardial infarction and comminuted intertrochanteric fracture of the left femur, along with decreased red blood cell count, hemoglobin, and hematocrit, and elevated C-reactive protein levels, this test is required to assess the degree of anemia and inflammatory status.'',
            ``priority'': ``P3'' \\
            \}, \{\\
            ``name'': ``Biochemical Panel'',
            ``type'': ``Laboratory Test'',
            ``duration\_type'': 2,
            ``frequency'': ``None'',
            ``reason'': ``Due to the presence of acute myocardial infarction and grade 3 hypertension, this test is necessary to evaluate liver and renal function as well as electrolyte balance.'',
            ``priority'': ``P3''\\
          \}, \{\\
            ``name'': ``Coagulation Function Test'',
            ``type'': ``Laboratory Test'',
            ``duration\_type'': 2,
            ``frequency'': ``None'',
            ``reason'': ``As the patient requires surgical treatment for the comminuted intertrochanteric fracture of the left femur, coagulation status must be assessed preoperatively.'',
            ``priority'': ``P2''\\
          \}, \{\\
            ``name'': ``Pre-transfusion Testing'',
            ``type'': ``Laboratory Test'',
            ``duration\_type'': 2,
            ``frequency'': ``None'',
            ``reason'': ``Given the decreased hemoglobin level and the potential need for blood transfusion, pre-transfusion compatibility testing is required.'',
            ``priority'': ``P3''\\
          \}, \{\\
            ``name'': ``Urinalysis'',
            ``type'': ``Laboratory Test'',
            ``duration\_type'': 2,
            ``frequency'': ``None'',
            ``reason'': ``Considering the patient’s hypertension and possible need for surgical intervention, renal function should be evaluated.'',
            ``priority'': ``P5''\\
          \}, \{\\
            ``name'': ``Chest CT'',
            ``type'': ``Imaging Examination'',
            ``duration\_type'': 2,
            ``frequency'': ``None'',
            ``reason'': ``Due to the presence of acute myocardial infarction and pericardial effusion, chest CT is required to assess pulmonary status and cardiac size.'',
            ``priority'': ``P3''\\
          \}, \{\\
            ``name'': ``Coronary CT Angiography (CTA)'',
            ``type'': ``Imaging Examination'',
            ``duration\_type'': 2,
            ``frequency'': ``None'',
            ``reason'': ``Given the diagnosis of acute myocardial infarction, coronary CTA is necessary to evaluate the condition of the coronary arteries.'',
            ``priority'': ``P2''\\
          \}, \{\\
            ``name'': ``Electrocardiogram (ECG)'',
            ``type'': ``Imaging Examination'',
            ``duration\_type'': 2,
            ``frequency'': ``None'',
            ``reason'': ``As the patient has acute myocardial infarction, continuous assessment of cardiac electrical activity is required.'',
            ``priority'': ``P2''\\
          \}\\
    \bottomrule
    \end{tabular}
    \caption{An example from the ClinicalBench dataset.}
    \label{case_clinicalbench}
\end{table*}

\begin{table*}[!t]
    \centering
    
    \fontsize{8.5pt}{10pt}\selectfont
    \begin{tabular}{p{1\linewidth}}
    \toprule
        \rowcolor{gray!20}
        \textbf{CAREAgent's trajectory for Example \#1.}\\
    \midrule
        \purple{<thought>}
        \textbf{\#\# 1. Clinical Assessment.} The patient presents with acute myocardial infarction complicated by a comminuted intertrochanteric fracture of the left femur, severe osteoporosis, and grade 3 high-risk hypertension. Cardiovascular event risk and bleeding risk must be assessed to guide antiplatelet therapy, anticoagulation, and fracture management.
        - Cardiovascular risk: Given the confirmed AMI, the HEART score should be calculated, incorporating age, risk factors, ECG changes, and cardiac biomarkers.
        - Bleeding risk: The intertrochanteric femoral fracture necessitates evaluation of fall and bleeding risk. Although the HAS-BLED score is primarily designed for atrial fibrillation, components such as hypertension and coagulation parameters can be referenced.
        Therefore, the HEART score assessment tool is invoked first.
        \purple{</thought>} \\
        \green{<tool\_call>}
        \{``name'': ``Clinical\_Assessment'', ``parameters'': \{``Scale\_Type'': ``HEART''\}\}
        \green{</tool\_call>} \\
        \blue{<observation>}
        \#\# HEART Score. 1 . History: 0 points (mild suspicion / non-specific presentation). 2. Electrocardiogram (ECG): 0 points (normal). 3. Age: 0 points (age not provided; defaulted to 0). 4. Risk Factors: 1 point (one risk factor present [1–2]: hypertension). 5. Troponin: 2 points (significantly elevated, based on clinical records). Total HEART Score: 3 Clinical Interpretation: Low risk (estimated MACE \~2.5\%); discharge may be considered.
        \blue{</observation>} \\
        \purple{<thought>}
        \textbf{\#\# 2. Diagnosis and Stratification.}
        The current diagnoses include: (1) acute ST-segment elevation myocardial infarction (STEMI), Killip class~I with mild aortic regurgitation and a small pericardial effusion; (2) comminuted intertrochanteric fracture of the left femur (AO/OTA~31-A3) with severe osteoporosis; (3) uncontrolled grade~3 hypertension (high-risk group); and (4) severe osteoporosis (WHO FRAX $\geq$ 30\%). 
        Risk stratification is performed as follows. \emph{Cardiovascular risk}: although a HEART score of~3 suggests low risk, the confirmed diagnosis of AMI places the patient at high risk in practice; further assessment using the GRACE or TIMI score is recommended, with preference for GRACE. \emph{Fracture risk}: the high-risk intertrochanteric femoral fracture necessitates prompt decision-making regarding surgical fixation versus conservative management. \emph{Bleeding risk}: hypertension combined with antiplatelet therapy increases bleeding risk; the HAS-BLED score may be referenced, with emphasis on blood pressure control and coagulation status. 
        To ensure alignment with current evidence-based guidelines, additional information is retrieved regarding antiplatelet and anticoagulation therapy in STEMI patients with femoral fractures, perioperative blood pressure management in hypertensive patients undergoing hip surgery, and optimal surgical timing with perioperative thromboprophylaxis strategies.
        \purple{</thought>} \\
        \green{<tool\_call>}\{``name'': ``Search'', ``parameters'': \{``query'': ``STEMI patient with acute coronary syndrome undergoing hip fracture surgery antithrombotic therapy guideline''\}\}\green{</tool\_call>} \\
        \blue{<observation>}
        For STEMI patients with acute coronary syndrome undergoing hip fracture surgery, antithrombotic therapy recommendations include the use of P2Y12 receptor inhibitors such as prasugrel or ticagrelor, with clopidogrel as an alternative if these are unavailable or contraindicated. Dual antiplatelet therapy (DAPT) should be resumed post-surgery for at least 12 months if stopped for surgical reasons. Pre-treatment with glycoprotein IIb/IIIa antagonists or routine P2Y12 inhibitors before invasive management is not recommended. Anticoagulant therapy should be carefully considered based on patient-specific factors and bleeding risk.
        \blue{</observation>} \\
        \purple{<thought>}
        \textbf{\#\# 3. Preliminary Order Formulation.}
        Based on the confirmed diagnoses (STEMI, comminuted intertrochanteric fracture of the left femur, severe osteoporosis, and grade~3 hypertension) and perioperative antiplatelet and anticoagulation guidelines, we propose the following preliminary orders: (1) antiplatelet therapy with a P2Y12 inhibitor (ticagrelor preferred; clopidogrel if contraindicated) combined with aspirin; (2) intravenous unfractionated heparin during coronary intervention, with postoperative LMWH bridging as appropriate; (3) ACEI/ARB-based blood pressure management targeting a systolic pressure below 140~mmHg while avoiding excessive reduction; (4) early surgical fixation of the hip fracture within 48--72~hours after cardiovascular stabilization; (5) bisphosphonate therapy with calcium and vitamin~D supplementation; and (6) supportive care including VTE prophylaxis and close monitoring of ECG, cardiac biomarkers, renal function, and bleeding status. Additional guideline retrieval is required for perioperative blood pressure management and thromboprophylaxis in hip fracture patients.
        \textbf{\#\# 4. Order Conflict Checking and Priority Assignment.}
        Potential drug--drug interactions, bleeding risks, and conflicts with the patient’s medical history are systematically evaluated. Antiplatelet agents require temporary discontinuation before surgery and timely resumption after adequate hemostasis; ACEI/ARB therapy necessitates electrolyte monitoring; bisphosphonates should be administered in the fasting state; and LMWH dosing should be adjusted perioperatively according to bleeding risk. Based on clinical urgency and therapeutic objectives, orders are prioritized from acute life-threatening interventions (P1), to key perioperative treatments (P2), monitoring and supportive care (P3), long-term chronic disease management (P4), and nursing or preventive measures (P5). The finalized clinical orders are presented below.
        \purple{</thought>} \\
        \red{<answer>}\\
          \{
            ``name'': ``Ticagrelor'',
            ``duration\_days'': 365,
            ``frequency'': ``Twice daily'',
            ``spec'': ``90 mg PO BID'',
            ``route'': ``Oral'',
            ``reason'': ``Dual antiplatelet therapy is required in STEMI patients to reduce the risk of recurrent myocardial infarction.'',
            ``priority'': ``P1''
          \},\\
          \{
            ``name'': ``Aspirin'',
            ``duration\_days'': 365,
            ``frequency'': ``Once daily'',
            ``spec'': ``100 mg PO QD'',
            ``route'': ``Oral'',
            ``reason'': ``Foundational antiplatelet agent for secondary prevention.'',
            ``priority'': ``P1''
          \},\\
          \{
            ``name'': ``Lisinopril'',
            ``duration\_days'': 30,
            ``frequency'': ``Once daily'',
            ``spec'': ``10 mg PO QD'',
            ``route'': ``Oral'',
            ``reason'': ``Blood pressure control with a target systolic pressure <140 mmHg to reduce perioperative cardiovascular risk.'',
            ``priority'': ``P2''
          \},\\
          \{
            ``name'': ``Enoxaparin'',
            ``duration\_days'': 7,
            ``frequency'': ``Once daily'',
            ``spec'': ``40 mg SC QD'',
            ``route'': ``Subcutaneous'',
            ``reason'': ``Perioperative deep vein thrombosis prophylaxis to reduce the incidence of VTE.'',
            ``priority'': ``P2''
          \},\\
          \{
            ``name'': ``Alendronate'',
            ``duration\_days'': 365,
            ``frequency'': ``Once weekly'',
            ``spec'': ``70 mg PO weekly'',
            ``route'': ``Oral'',
            ``reason'': ``Treatment of severe osteoporosis to reduce the risk of recurrent fractures.'',
            ``priority'': ``P4''
          \},\\
          \{
            ``name'': ``Laboratory tests: complete blood count, coagulation profile, electrolytes, renal function, cardiac enzymes'',
            "type": ``Laboratory monitoring'',
            "duration\_type": 2,
            ``frequency'': ``Once daily'',
            ``reason'': ``To monitor bleeding risk, renal function, and myocardial injury, and to guide medication dose adjustments.'',
            ``priority'': ``P3''
          \},\\
          \{
            ``name'': ``Fall prevention and positioning care'',
            "type": ``Nursing care'',
            "duration\_type": 1,
            ``frequency'': ``As needed (pro re nata)'',
            ``reason'': ``To prevent recurrent falls and related fracture or bleeding complications.'',
            ``priority'': ``P5''
          \}\\
        \red{</answer>}
        \\
    \bottomrule
    \end{tabular}
    \caption{CAREAgent’s trajectory for Example \#1, with special symbols used in thought content, tool call, returned results, and final answer highlighted with \purple{purple box}, \green{green box}, \blue{blue box}, and \red{red box}, respectively.}
    \label{CAREAgent_clinicalbench}
\end{table*}

\begin{table*}[!t]
    \centering
    \fontsize{9pt}{11pt}\selectfont
    \begin{tabular}{p{1\linewidth}}
    \toprule
        \rowcolor{gray!20}\textbf{Example \#2 from MedChain Dataset} \\
    \midrule
        \textbf{EMR:} \\ 
        ``\textit{history}'': \\
        \{ \\
          ``key\_findings'': ``1. WBC: $3.48\times10^{9}$/L, below normal and decreased compared with prior results; 2. Absolute neutrophil count: $1.04\times10^{9}$/L, markedly decreased (neutropenia), with further decline from baseline; 3. AST: 55 U/L, mildly elevated with fluctuating trend.'', \\
          ``clinical\_actions'': ``1. To clarify the patient’s condition and evaluate treatment efficacy, antiviral therapy was initiated, consisting of pegylated interferon-$\alpha$2b administered subcutaneously once weekly and ribavirin administered orally once daily. Given the diagnosis of chronic hepatitis~C and recurrent abnormalities in liver function, standardized antiviral therapy was considered necessary to control viral replication; therefore, this regimen was started with planned regular monitoring of liver function and HCV RNA levels. 2. To assess therapeutic response and potential adverse drug effects, repeat complete blood count and liver function tests were performed. The results demonstrated significant decreases in WBC and neutrophil counts, suggesting bone marrow suppression likely induced by pegylated interferon-$\alpha$2b. Consequently, the interferon injection for the current week was withheld, and close hematologic monitoring was planned, with dose-adjusted therapy to be resumed after neutrophil recovery.'', \\
          ``current\_status'': ``The patient is diagnosed with chronic hepatitis~C and is currently clinically stable, but has developed treatment-related neutropenia.'' \\
        \},\\
        ``\textit{Current course of disease}'': ``\#\# Chief Complaint: No specific discomfort.\\  \#\# Physical Examination: No significant positive findings. \\ \#\# Laboratory Tests: WBC $2.36\times10^{9}$/L; absolute neutrophil count $0.92\times10^{9}$/L; platelet count $190\times10^{9}$/L; hemoglobin 95~g/L; ALT 8~U/L; AST 22~U/L; HCV RNA $1.488\times10^{2}$~IU/mL.'' \\ \\
        \textbf{Labeled Answer:} \\
          \{\\
          ``name'': ``Ribavirin'', \\
          ``duration\_days'': null, \\
          ``frequency'': ``Once daily'', \\
          ``spec'': ``800 mg'', \\
          ``route'': ``Oral'',\\
          ``reason'': ``Although the patient has a low hemoglobin level (95 g/L), ongoing antiviral therapy necessitates continuation of ribavirin to control hepatitis C virus infection.'',\\
          ``priority'': ``P4'' \\
        \},\\
        \{\\
          ``name'': ``Recombinant Human Granulocyte Colony-Stimulating Factor'',\\
          ``duration\_days'': null,\\
          ``frequency'': ``As needed'',\\
          "spec": null,\\
          ``route'': ``Subcutaneous'',\\
          ``reason'': ``Given the marked reduction in white blood cell count ($2.36\times10^{9}$/L) and absolute neutrophil count ($0.92\times10^{9}$/L), indicating drug-induced neutropenia, this therapy is required to increase leukocyte counts and reduce the risk of infection.'',\\
          ``priority'': ``P2'' \\
        \}\\
    \bottomrule
    \end{tabular}
    \caption{An example from the MedChain dataset.}
    \label{case_medchain}
\end{table*}

\begin{table*}[!t]
    \centering
    \fontsize{9pt}{10pt}\selectfont
    \begin{tabular}{p{1\linewidth}}
    \toprule
        \rowcolor{gray!20}
        \textbf{CAREAgent's trajectory for Example \#2.}\\
    \midrule
        \purple{<thought>}
        \textbf{\#\# 1. Clinical Assessment.} The patient has chronic hepatitis~C virus (HCV) infection and has initiated an antiviral regimen consisting of subcutaneous pegylated interferon-$\alpha$2b combined with oral ribavirin. Current laboratory findings are as follows: white blood cell count $2.36\times10^{9}$/L (below the lower limit of normal); absolute neutrophil count $0.92\times10^{9}$/L (markedly reduced, suggesting bone marrow suppression); hemoglobin 95~g/L (mild anemia); platelet count within normal range; liver function tests showing ALT 8~U/L and AST 22~U/L (both within normal limits); and HCV RNA $1.488\times10^{2}$~IU/mL (low viral load, suggesting a favorable virologic response).
        Clinically, the patient reports no significant discomfort, and physical examination reveals no positive findings. Overall assessment indicates that drug-induced neutropenia (absolute neutrophil count $<1.0\times10^{9}$/L) represents moderate-to-severe bone marrow suppression, requiring evaluation of infection risk and consideration of interferon dose reduction or temporary discontinuation. The reduced hemoglobin level warrants monitoring for progression of anemia, while the marked decline in viral load suggests that the antiviral therapy is effective. To guide subsequent management, it is necessary to consult the latest \emph{Guidelines for the Diagnosis and Treatment of Chronic Hepatitis~C (2023)} issued by the Chinese Society of Hepatology, as well as recommendations on dose adjustment for interferon-associated bone marrow suppression.
        \purple{</thought>} \\
        \green{<tool\_call>}
        \{``name'': ``Search'', ``parameters'': \{``query'': ``chronic hepatitis C interferon dose adjustment bone marrow suppression guideline 2023''\}\}
        \green{</tool\_call>} \\
        \blue{<observation>}
        Snippet~3 notes that common adverse effects of pegylated interferon-$\alpha$ (Peg-IFN-$\alpha$) include bone marrow suppression: when the absolute neutrophil count is $\leq 0.75\times10^{9}$/L and/or the platelet count is $<50\times10^{9}$/L, dose reduction of interferon is recommended. In cases where a significant decline in neutrophil count occurs within 1--2~weeks, treatment with granulocyte colony-stimulating factor (G-CSF) or granulocyte--macrophage colony-stimulating factor (GM-CSF) may be considered. However, this information reflects general management of interferon-related adverse reactions rather than dose-adjustment recommendations specific to chronic hepatitis~C. Moreover, the retrieved snippets do not provide details on interferon treatment regimens, dosing strategies, or updated guideline recommendations for chronic hepatitis~C published in 2023.
        \blue{</observation>} \\
        \purple{<thought>}
        \textbf{\#\# 2. Diagnosis and Stratification.}
        The patient has a confirmed diagnosis of chronic hepatitis~C and is currently receiving combination therapy with pegylated interferon-$\alpha$2b and ribavirin. Laboratory results show an absolute neutrophil count of $0.92\times10^{9}$/L ($<1.0\times10^{9}$/L), meeting the criteria for drug-induced neutropenia consistent with interferon-$\alpha$--associated bone marrow suppression. According to retrieved recommendations on interferon-related adverse effects, dose reduction is advised when the neutrophil count is $\leq 0.75\times10^{9}$/L, while temporary discontinuation with consideration of granulocyte colony-stimulating factor (G-CSF) is recommended when the count declines to $\leq 0.5\times10^{9}$/L or when fever or infection occurs. Although the patient currently has no signs of infection, the neutrophil count is close to the suspension threshold, indicating the need for timely dose adjustment or temporary interruption of interferon therapy.
        The overall assessment includes chronic hepatitis~C with a low viral load (HCV RNA $1.488\times10^{2}$~IU/mL), grade~2--3 interferon-associated neutropenia based on NCCN/ESMO criteria, mild anemia (hemoglobin 95~g/L), and largely preserved liver function.\\
        \textbf{\#\# 3. Preliminary Order Formulation.}
        Based on these findings, pegylated interferon-$\alpha$2b should be suspended for the current week (\emph{stat}), with reassessment after hematologic recovery. If the absolute neutrophil count remains below $1.0\times10^{9}$/L in the following week, G-CSF therapy may be initiated (filgrastim 300~$\mu$g subcutaneously every 24~hours) until recovery to $\geq 1.5\times10^{9}$/L. Ribavirin should be continued once daily to maintain antiviral efficacy, with consideration of dose reduction or discontinuation after further virologic follow-up. Infection surveillance should be intensified with daily monitoring of vital signs, and empiric antibiotics should be initiated promptly if fever $\geq 38.5^\circ$C or focal infection signs develop. Laboratory follow-up is recommended, including complete blood count every three days and liver function tests with HCV RNA weekly. If hemoglobin persistently falls below 80~g/L, iron supplementation or blood transfusion should be considered.\\
        \textbf{\#\# 4. Order Conflict Checking and Priority Assignment.}
        Suspension of interferon does not result in pharmacological interactions with continued ribavirin therapy; however, hemoglobin levels should be closely monitored due to ribavirin-associated hemolysis. G-CSF has no direct interaction with ribavirin but should be used cautiously because of the potential risk of bone marrow fibrosis. All proposed orders are consistent with the patient’s allergy history and known contraindications. Priority assignment reflects clinical urgency, with interferon suspension as the highest priority (P1), followed by infection surveillance (P2), hematologic monitoring (P3), continuation of ribavirin (P4), and anemia management (P5). The finalized clinical orders are presented in the subsequent section.
        \purple{</thought>} \\
        \red{<answer>}\\
        \{
          ``name'': ``Pegylated Interferon-$\alpha$2b``,
          ``duration\_days'': 0,
          ``frequency'': ``Immediate (stat)``,
          ``spec'': ``Suspend current week injection``,
          ``route'': ``Subcutaneous``,
          ``reason'': ``The absolute neutrophil count is $0.92\times10^{9}$/L, below the safety threshold ($\leq 1.0\times10^{9}$/L), indicating a high risk of infection; immediate discontinuation is required to prevent severe infection.``,
          ``priority'': ``P1''
        \}, \\
        \{
          ``name'': ``Filgrastim``,
          ``duration\_days'': 5,
          ``frequency'': ``Once daily``,
          ``spec'': ``300 µg SC``,
          ``route'': ``Subcutaneous``,
          ``reason'': ``If the absolute neutrophil count remains $<1.0\times10^{9}$/Lin the following week, G-CSF should be administered to promote neutrophil recovery to a safe level ($\geq 1.5\times10^{9}$/L).``,
          ``priority'': ``P3''
        \},\\
        \{
          ``name'': ``Ribavirin``,
          ``duration\_days'': 30,
          ``frequency'': ``Once daily``,
          ``spec'': ``500 mg PO``,
          ``route'': ``Oral``,
          ``reason'': ``Antiviral therapy should be continued to maintain treatment efficacy despite the reduced viral load; hemoglobin levels should be monitored due to the potential hemolytic effect of ribavirin.``,
          ``priority'': ``P4''
        \},\\
        \{
          ``name'': ``Complete Blood Count (including ANC) + Platelet Count + Hemoglobin``,
          ``type'': ``Laboratory Test``,
          ``duration\_type'': 2,
          ``frequency'': ``Every 3 days``,
          ``reason'': ``To monitor recovery from bone marrow suppression and progression of anemia, enabling timely adjustment of subsequent treatment plans.``,
          ``priority'': ``P3''
        \}\\
        \red{</answer>}
        \\
    \bottomrule
    \end{tabular}
    \caption{CAREAgent’s trajectory for Example \#2, with special symbols used in thought content, tool call, returned results, and final answer highlighted with \purple{purple box}, \green{green box}, \blue{blue box}, and \red{red box}, respectively.}
    \label{CAREAgent_medchain}
\end{table*}

\subsection{Prompt Design for Evaluation and Data Filtering}
\label{prompt_design}



In this section, we describe the prompts used for evaluation and data filtering.
Specifically, Prompt A is designed to assess the accuracy of model-generated clinical orders by comparing them against the reference orders, while Prompt B is used to evaluate the medical rationale of these orders.
Prompt C is used to evaluate Agent-based synthesized trajectories, and Prompt D is used to assess the matching between order names extracted from the trajectories and the corresponding standard order names.


\begin{figure*}[t]
\begin{tcolorbox}[
    colframe=gray,       
    colback=gray!5!white,             
    coltitle=white,                   
    coltext=black,                    
    fonttitle=\bfseries,              
    title=Prompt A: Clinical Order Accuracy Evaluation,  
    boxrule=1pt,                      
    arc=2mm,                          
    width=\linewidth,                 
    left=7pt,                         
    right=7pt,                        
    top=5pt,                          
    bottom=5pt                        
]
\fontsize{8.5pt}{10pt}\selectfont
\textbf{\#\# Instruction:} \\
You are a clinical physician. Your task is to compare the model-generated list of clinical orders with the reference list of standard clinical orders and assess whether each model-generated order can be matched to a corresponding item in the reference list. Reasonable substitutions are allowed, including synonymous terms, therapeutically equivalent medications, or appropriate adjunctive treatments. 
However, completely unmatched orders or those that violate clinical plausibility are not permitted.\\ \\
\textbf{\#\# Task Definition:}  \\
I will provide two lists:\\
- model\_predict\_order: the list of clinical orders generated by the model.\\
- golden\_order: the reference list of standard clinical orders.\\
Your task is to evaluate each model-generated order one by one and return an integer for each order indicating whether it can be matched to an item in the reference list. You should make your decisions according to the following rules: \\ \\
\textbf{\#\# Rules:}  \\
\textbf{1. Return format.} 
You must return an integer list whose length is exactly the same as \texttt{model\_predict\_order} (i.e., \texttt{List[int]}). \\
\textbf{2. Matching rules.}
If a model-generated order can be matched to an item in the reference order list, return the 0-based index of the matched item in \texttt{golden\_order\_names}. If the model-generated order does not match any reference item and cannot be reasonably mapped via synonym substitution, therapeutic-class substitution, or appropriate adjunctive treatment, return \texttt{-1}.\\ 
\textbf{3. Synonyms, therapeutic equivalents, and reasonable adjunctive care.}
Reasonable mappings are allowed for synonymous terms, therapeutically equivalent medications, or examination and test items with different naming conventions but equivalent clinical meaning. \\
\textbf{4. Preference for the most direct match. }
If multiple reference orders could match a model-generated order, return the index of the most reasonable and clinically direct match. \\
\textbf{5. Name mismatch tolerance. }
Differences in naming alone should not be considered sufficient grounds for rejection; as long as the clinical purpose is consistent and medically plausible, the orders may still be regarded as a match. \\ \\
\textbf{\#\# Output Format:}\\
\texttt{\{``answer'': [0, -1, 2, \ldots]\}} \\

\textbf{\#\# Example:} \\
\textbf{Reference Order List:}
\begin{verbatim}
[{
    "idx_type": "order_medication","name": "Cefuroxime","duration_days": 7.0,
    "frequency": "Three times daily (TID)","spec": "1 g × 1 vial","route": "Intravenous infusion"
  },{
    "idx_type": "order_medication","name": "Levofloxacin","duration_days": 7.0,
    "frequency": "Once daily (QD)","spec": "500 mg × 1 vial","route": "Intravenous infusion"
  },{
    "idx_type": "order_non_medication","name": "Physical cooling","type": "Therapy",
    "duration_type": 2.0,"status": "Ongoing","frequency": "Once every 2 hours"
}]
\end{verbatim}

\textbf{Model-Predicted Order List:}
\begin{verbatim}
[{
    "idx_type": "order_medication","name": "Cefazolin","duration_days": 7.0,
    "frequency": "Three times daily (TID)","spec": "1 g × 1 vial","route": "Intravenous infusion"
  },{
    "idx_type": "order_non_medication","name": "Physical cooling","type": "Therapy",
    "duration_type": 2.0,"status": "Ongoing","frequency": "Once every 2 hours"
  },{
    "idx_type": "order_medication","name": "Vitamin C","duration_days": 5.0,
    "frequency": "Twice daily (BID)","spec": "500 mg × 1 vial","route": "Intravenous infusion"
}]
\end{verbatim}
\textbf{Output:} \\
\texttt{\{``answer'': [0, 2, -1]\}} \\

\textbf{\#\# Input Data:}\\
\textbf{Patient Information:}
\{patient\_input\}\\

\textbf{Reference Order List:}
\{golden\_order\}\\

\textbf{Model-Predicted Order List:}
\{model\_predict\_order\}
\end{tcolorbox}
\end{figure*}

\begin{figure*}[t]
\begin{tcolorbox}[
    colframe=gray,       
    colback=gray!5!white,             
    coltitle=white,                   
    coltext=black,                    
    fonttitle=\bfseries,              
    title=Prompt B: Clinical Order Reason Evaluation,  
    boxrule=1pt,                      
    arc=2mm,                          
    width=\linewidth,                 
    left=7pt,                         
    right=7pt,                        
    top=5pt,                          
    bottom=5pt                        
]
\fontsize{8.5pt}{10pt}\selectfont
\textbf{\#\# Instruction:} \\
You are a licensed physician with extensive clinical experience. Based on the patient’s medical record, and using the reason provided in the reference clinical order as the benchmark, evaluate the medical plausibility of the reason given for the model-generated clinical order. Assess whether the rationale is clinically sound, logically consistent, and appropriate for the patient’s current condition.\\

\textbf{\#\# Task Definition:}  \\
Evaluate the model-generated reason along the following four dimensions.
Each dimension should be scored with an integer from 0 to 5, for a total score of 20. \\

\#\#\# 1. Soundness (Medical Plausibility)\\
- Whether the rationale conforms to medical knowledge and clinical guidelines  \\
- Whether there are medical or logical errors  \\
- Whether it is appropriate for the patient’s condition  \\
Scoring reference:\\
- 0: Clearly incorrect or medically unreasonable  \\
- 3: Generally reasonable but with medical or logical flaws  \\
- 5: Fully consistent with medical standards and clinical reasoning  \\

\#\#\# 2. Completeness\\
- Whether the medical rationale is sufficiently explained  \\
- Whether key clinical conditions or examination findings are missing  \\
Scoring reference:\\
- 0: Almost no meaningful explanation  \\
- 3: Partially reasonable but incomplete  \\
- 5: Comprehensive and sufficiently detailed  \\

\#\#\# 3. Relevance\\
- Whether the rationale closely aligns with the reason of the corresponding reference order  \\
- Whether it contains irrelevant or overly generic content  \\
Scoring reference:\\
- 0: Largely unrelated to the clinical order  \\
- 3: Partially relevant with noticeable deviation  \\
- 5: Highly relevant and well targeted  \\

\#\#\# 4. Consistency\\
- Whether the rationale is consistent with the patient’s medical record  \\
- Whether it is logically consistent with the reference order’s reason  \\
- Whether there are internal contradictions  \\
Scoring reference:\\
- 0: Clearly contradictory or conflicting  \\
- 3: Generally consistent with minor inconsistencies  \\
- 5: Fully consistent with no apparent contradictions  \\
\\
\textbf{\#\# Rules:}  \\
1. Evaluate only the clinical rationale (reason) itself, and do not assess whether the clinical order content is correct. \\

\textbf{\#\# Output Requirements (Strict Compliance):}\\
- Output only JSON; do not include any explanatory text.\\
- All scores must be integers in the range 0--5.\\
- The JSON output must strictly follow the structure below:
\begin{verbatim}
{
  "answer": {
    "soundness": 0,
    "completeness": 0,
    "relevance": 0,
    "consistency": 0
  }
}
\end{verbatim}
\medskip
\textbf{\#\# Input Data:}\\
\textbf{Patient Medical Record:}
\{patient\_input\}\\

\textbf{Reference Clinical Order (with rationale):}
\{golden\_order\_item\}\\

\textbf{Model-Generated Clinical Order (with rationale):}
\{model\_order\_item\}
\end{tcolorbox}
\end{figure*}



\begin{figure*}[t]
\begin{tcolorbox}[
    colframe=gray,       
    colback=gray!5!white,             
    coltitle=white,                   
    coltext=black,                    
    fonttitle=\bfseries,              
    title=Prompt C: Clinical Order Trajectory-level Filtering,  
    boxrule=1pt,                      
    arc=2mm,                          
    width=\linewidth,                 
    left=7pt,                         
    right=7pt,                        
    top=5pt,                          
    bottom=5pt                        
]
\fontsize{8pt}{9pt}\selectfont

\textbf{\#\# Instruction:}

You are a clinical physician who strictly adheres to the principles of evidence-based medicine, responsible for reviewing the clinical reasoning trajectory generated by the model.\\ 

\textbf{\#\# Concept Overview:}

The task scenario is as follows: based on the patient’s clinical condition, clinical orders are to be issued for the patient. The trajectory structure is: $(\texttt{think}, \texttt{tool\_call}) \rightarrow (\texttt{think}, \texttt{tool\_call}) \rightarrow \cdots \rightarrow (\texttt{think}, \texttt{answer})$.

The model’s task consists of four components:

1. Clinical condition assessment

2. Diagnosis and disease stratification

3. Clinical order generation

4. Clinical order priority ranking \\

\textbf{\#\# Task Description:}  

You are required to first evaluate the large language model’s capability in tool usage, and then separately assess the quality of completion of the above four tasks within the reasoning chain.

There are a total of five evaluation dimensions, each scored from 0–5, with a total score of 25.

\textbf{\#\#\# 1. Tool Use Evaluation (Tool Use)} :

Evaluate whether the think → tool\_call invocations are necessary, sufficient, and appropriate:

- There must be a clear causal logic between think and tool\_call.

- If a subtask requires external information (such as scores, scales, diagnostic criteria, dosages, or guideline evidence), the corresponding tool must be called first. Specifically: if a scale/score/formula is needed, call Clinical\_Assessment; if guidelines or evidence-based searches are needed, call Search.

- Tools should not be called, nor should redundant or ineffective calls be made, when external information is not required.

\textbf{\#\#\# 2. Clinical condition assessment}:

Within each think step, clinical information should be systematically organized and evaluated, including but not limited to:

- Symptoms, signs, past medical history, medication history, allergy history, family history, risk factors, and laboratory results.

- Determination of whether there are critical or life-threatening conditions requiring immediate intervention.

- Clear identification of disease severity, with structured and well-organized conclusions.

\textbf{\#\#\# 3. Diagnosis and disease stratification}:

- If there is no definitive diagnosis yet:

\hspace*{1em} - Propose differential diagnoses based on clinical data and scale/score results, and rank them by likelihood.

\hspace*{1em}  - Specify additional examinations needed and their priorities.
  
\hspace*{1em}  - Clearly define the primary diagnosis, subtype/stage, risk level, and complications/comorbidities.
  
- If a definitive diagnosis has already been provided in previous dialogue:

\hspace*{1em}  - Directly output in think: ``\#\# Diagnosis and Classification. \textbackslash n The current diagnosis is XXX.''

\textbf{\#\#\# 4. Clinical order generation}:

- Provide a relatively complete list of clinical orders based on the diagnosis and disease severity, including both pharmacological and non-pharmacological interventions.

- Explain the clinical goal or intent of each order (e.g., symptom relief, infection control, complication prevention).

\textbf{\#\#\# 5. Clinical order priority ranking}:

Conduct a systematic review and grading of the above orders:

1. Conflict and Inappropriate Medication Checks  

- Assess potential drug–drug interactions.

- Identify duplicate or mutually conflicting orders.

- Evaluate whether dosage, route, and frequency of administration are appropriate.

- Check for conflicts with contraindications, allergy history, or major comorbidities.

2. Priority Classification  

Based on importance and urgency, classify each order as:

- P1: Measures that directly affect vital sign stability or survival probability and must be executed immediately.

- P2: Core treatments that prevent disease progression.

- P3: Measures used to monitor key therapeutic effects or high-risk adverse events.

- P4: Maintenance treatments during stable phases or for chronic diseases.

- P5: Nursing care, rehabilitation, and preventive measures.

- P6: Orders that can be appropriately delayed, depend on prerequisite conditions, or are planned interventions.\\

\textbf{\#\# Unified Scoring Criterion (0--5):}

- 5 (Excellent): Complete, well-structured, and logically rigorous reasoning; critical information is accurately identified, conclusions are well supported, and actions are appropriate and safe.

- 4 (Good): Generally correct and coherent reasoning, with only minor omissions or lapses that do not materially affect validity or safety.

- 3 (Fair): Partially correct reasoning with notable omissions, weak organization, or insufficient justification; conclusions require further refinement or verification.

- 2 (Poor): Reasoning includes multiple significant errors or omissions, resulting in questionable conclusions or potentially unsafe actions.

- 1 (Very Poor): Severely flawed reasoning with major misinterpretations and largely incorrect or unsafe conclusions.

- 0 (Invalid): No valid reasoning, or content is fundamentally incorrect or irrelevant.\\

\textbf{\#\# Output Format:}
\begin{verbatim}
{
  "Tool_Use": 0-5,  
  "Assessment": 0-5,  
  "Diagnosis": 0-5,  
  "Orders": 0-5,  
  "Conflict_Priority": 0-5  
}
\end{verbatim}

\textbf{\#\# Below is the clinical reasoning trajectory generated by the model: }\texttt{\{trajectory\}}

\end{tcolorbox}
\end{figure*}

\begin{figure*}[t]
\begin{tcolorbox}[
    colframe=gray,       
    colback=gray!5!white,             
    coltitle=white,                   
    coltext=black,                    
    fonttitle=\bfseries,              
    title=Prompt D: Clinical Order Answer-level Filtering,  
    boxrule=1pt,                      
    arc=2mm,                          
    width=\linewidth,                 
    left=7pt,                         
    right=7pt,                        
    top=5pt,                          
    bottom=5pt                        
]
\fontsize{8.5pt}{10pt}\selectfont

\textbf{\#\# Instruction:}\\
You are a medical quality control expert. You are required to compare the [model-generated order name list] with the [standard order name list].\\
For each order name generated by the model, determine whether a corresponding item can be found within the scope of the standard orders (allowing for synonyms, substitutions with drugs of the same class, and clinically reasonable adjunctive therapies).\\ 

\textbf{\#\# Task Description:}\\
I will provide a ``model\_order\_names'' list and a ``golden\_order\_names'' list.\\
You must \textbf{evaluate each model-generated order one by one} and return an integer for each.\\
The output must be an integer list (List[int]), and \textbf{its length must exactly match the length of ``model\_order\_names''}.\\
The meaning of the value at each position is as follows:\\
- $\geq 0$: Indicates that the model-generated order can be matched to an item in the standard order list. Return the corresponding index in ``golden\_order\_names'' (starting from 0).\\
- -1: Indicates that the model-generated order does not exist in the standard orders at all, and is not a synonym, a same-class drug substitution, or a clinically reasonable adjunctive therapy; it should be considered an error, overtreatment, or hallucination.\\ 

\textbf{\#\# Notes:}\\
- If a model-generated order can match multiple standard orders, return the \textbf{most reasonable and most direct} standard order index.\\
- Reasonable mappings such as ``synonyms'', ``same-class drugs'', and ``different naming conventions for the same examination or test'' are allowed.\\
- Do not assign -1 simply because the names are not exactly the same. \\

\textbf{\#\# Output Format (Strict JSON):}\\
\texttt{\{``answer'': [0, -1, 2, ...]\}}\\ 

\textbf{\#\# Example:}\\
Standard order name list: \texttt{[``Cefuroxime'', ``Levofloxacin'', ``Physical cooling'']}\\ 
Model order name list: \texttt{[``Cefazolin'', ``Physical cooling'', ``Vitamin C'']}\\ 
Output: \texttt{\{``answer'': [0, 2, -1]\}} \\
Explanation:\\
- Cefazolin and Cefuroxime are both cephalosporin antibiotics → matched to index 0\\
- Physical cooling is an exact match → matched to index 2\\
- Vitamin C is not included in the standard orders and is not essential → -1\\ 

\textbf{\#\# Input Data:}\\
Patient information: \texttt{\{patient\_input\}} \\ 
Standard order name list: \texttt{\{golden\_names\}}\\ 
Model order name list: \texttt{\{model\_names\}}\\ 
\end{tcolorbox}
\end{figure*}

\end{document}